\documentclass{ieeetj}
\usepackage{amsmath,amssymb,amsfonts}
\usepackage{algorithmic}
\usepackage{graphicx,color}
\usepackage{textcomp}
\usepackage{xcolor}
\usepackage{hyperref}
\usepackage{soul}
\usepackage{algorithm}

\DeclareMathOperator*{\argmin}{arg\!\min}
\usepackage{dsfont}
\usepackage[
    backend=bibtex8,
    style=ieee,
    sorting=none,
    natbib=true,
    doi=false,
    isbn=false,
    url=false,
    eprint=false,
    maxcitenames=1,
    mincitenames=1,
    maxbibnames=20
]{biblatex}
\renewcommand{\sl}[1]{\textcolor{olive}{SL: {#1}}}

\usepackage{multirow}
\usepackage{array}
\usepackage{tabularx}
\usepackage{boldline}
\newcolumntype{q}[1]{>{\raggedleft\arraybackslash}p{#1}}
\newcolumntype{o}[1]{>{\centering\arraybackslash}p{#1}}
\newcolumntype{R}{>{\raggedleft\arraybackslash}X}
\newcolumntype{C}{>{\centering\arraybackslash}X}

\usepackage[printonlyused]{acronym}

\acrodef{LUV}{Lunar Utility Vehicle}
\acrodef{LTV}{Lunar Terrain Vehicle}
\acrodef{INS}{inertial navigation system}
\acrodef{GNSS}{global navigation satellite system}

\acrodef{GCS}{ground control station}
\acrodef{NTP}{network time protocol}

\acrodef{MPC}{model-predictive controller}
\acrodef{D-MPC}{distributed model-predictive-controller}

\acrodef{RMSE}{root-mean-squared error}

\acrodef{UTIAS}{University of Toronto Institute for Aerospace Studies}

\newcommand{\LELR}[0]{\ifx\blind\undefined LELR\else The rover\fi}

\newcommand{\lelr}[0]{\ifx\blind\undefined LELR\else the rover\fi}

\def\BibTeX{{\rm B\kern-.05em{\sc i\kern-.025em b}\kern-.08em
    T\kern-.1667em\lower.7ex\hbox{E}\kern-.125emX}}
\AtBeginDocument{\definecolor{tmlcncolor}{cmyk}{0.93,0.59,0.15,0.02}\definecolor{NavyBlue}{RGB}{0,86,125}}

\def\OJlogo{\vspace{-4pt}$<$Society logo(s) and publication title will appear here.$>$}
\def\seclogo{\vspace{10pt}$<$Society logo(s) and publication title will appear here.$>$}

\def\authorrefmark#1{\ensuremath{^{\textbf{#1}}}}

\begin{document}

\receiveddate{XX Month, XXXX}
\reviseddate{XX Month, XXXX}
\accepteddate{XX Month, XXXX}
\publisheddate{XX Month, XXXX}
\currentdate{XX Month, XXXX}
\doiinfo{XXXX.2022.1234567}

\markboth{}{Krawciw {et al.}}

\title{Sharing the Load: Autonomous Multi-Rover Cargo Transport}

\author{Alexander Krawciw\authorrefmark{1}, Graduate Student Member, IEEE, Luka Antonyshyn\authorrefmark{1},\\Sven Lilge\authorrefmark{1,2}, Nicolas Olmedo\authorrefmark{3}, Faizan Rehmatullah\authorrefmark{3} , Maxime Desjardins-Goulet\authorrefmark{4} , Pascal Toupin\authorrefmark{4}, and Timothy D. Barfoot\authorrefmark{1}, Fellow, IEEE}
\affil{University of Toronto Institute for Aerospace Studies, Toronto ON, M5S 1A1, Canada}
\affil{Toronto Metropolitan University, Toronto, ON, M5B 2K3, Canada}
\affil{MDA Space, Brampton, QC, M5S 1A1, Canada}
\affil{Centre de Technologies Avancées, Sherbrooke, QC J1K 0A5, Canada}
\corresp{Corresponding author: Alexander Krawciw (email: alec.krawciw@mail.utoronto.ca).}
\authornote{This work was supported in part by the Canadian Space Agency and the Natural Sciences and Engineering Research Council of Canada.}

\begin{abstract}
A future lunar habitat, as part of the Artemis program, will require a significant amount of logistics infrastructure. 
Cargo that is transported to the Moon will need to be moved from a landing site to other key locations that may be up to 5 km away. 
Teach and repeat navigation is well suited to this task as utility rovers will need to repeat these cargo routes many times. 
One of the most significant challenges involves the modules that will be assembled together to form the habitat. 
Canada is studying potential \ac{LUV} designs to carry these large payloads between the landing site and the location of the habitat.
As the details of the cargo continue to evolve, using two, smaller \acp{LUV} to carry cargo together would provide high capacity and mission flexibility. 
In this paper, we develop and implement a distributed model-predictive controller that allows vehicles to carry cargo that is shared between them. 
The algorithm is compared to baselines in small-scale before being implemented onboard two 800 kg path-to-flight rovers and field tested carrying a 475 kg cargo between them. 
A custom cargo coupling decouples the kinematics of each vehicle while fully supporting the cargo's mass. 
In our field test, the rovers maintain a relative separation error of 9.2 cm and maximum error of 33.4 cm. 
This multi-vehicle control architecture retains the high-quality path tracking of lidar teach and repeat for each rover. 
We demonstrate that kinematic freedom of the vehicles allows a single controller to provide mission improvements for other operations as well. 
\end{abstract}


\maketitle

\section{INTRODUCTION}
\begin{figure}
  \centering
  \includegraphics[width=\linewidth]{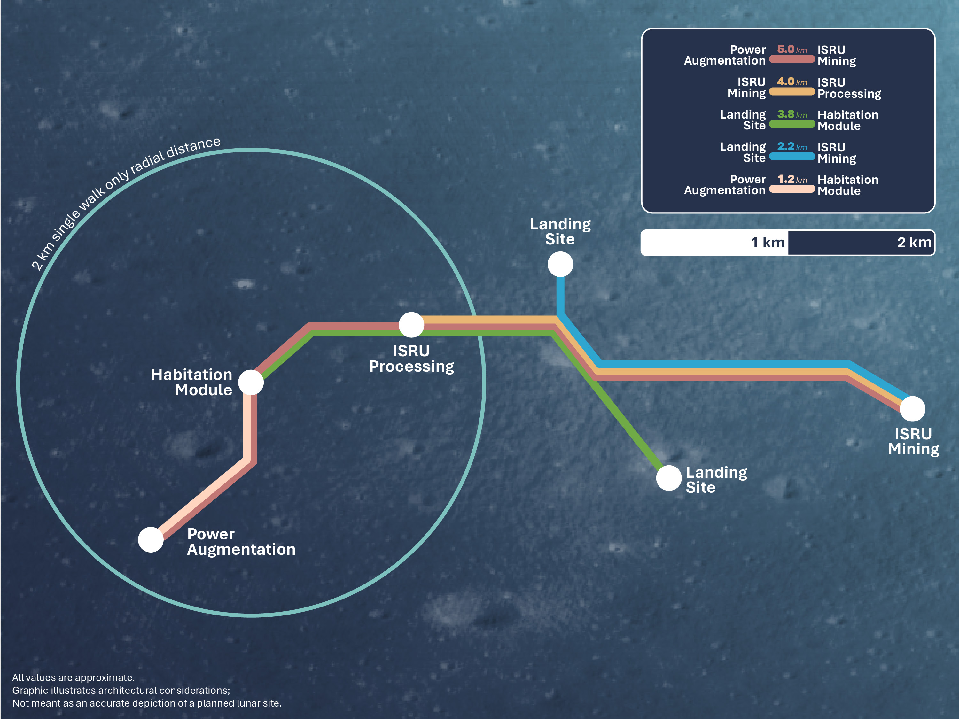}
  \vspace{-3mm}
  \caption{A conceptual layout of the different locations that cargo needs to be transported to. The network of routes would be autonomously navigated many times using a teach-and-repeat approach. Figure Credit: NASA \cite{nasa_2024a}.}
  \label{fig:path-network}
  \vspace{-5mm}
\end{figure}

\acresetall
\IEEEPARstart{C}{arrying} large pieces of cargo on the lunar surface is an open problem that is rapidly becoming more relevant. 
NASA's Artemis program aims to establish a semi-permanent habitat for astronauts to live in while performing lunar research \cite{nasa_2020}. 
Current concepts feature large pressurized modules that dock to each other, similar to the International Space Station, to create a larger habitable unit. 
These prefabricated modules will be transported to the moon on a large lander. 
However, once reaching the lunar surface, these modules will need to be moved up to 5 km \cite{nasa_2024} from the landing site to the habitat assembly location.
Current concepts call for a large automated vehicle, the \ac{LUV}, to move these pieces of cargo \cite{csa_2023}.
\autoref{fig:path-network} shows a conceptual arrangement of some of the key Artemis locations including landing sites, a habitat, and in-situ resource utilization (ISRU) regions. 
Transporting cargo along a common network of routes in unstructured, \ac{GNSS}-denied environments is a task well-suited to teach-and-repeat navigation \cite{krawciw_2025a}.
In teach and repeat, the vehicle is piloted along a network of paths using either teleoperation or a slower autonomy stack to map the environment and teach the routes. 
Subsequently, this map can be used for precise autonomous driving at higher speeds.
By driving along precisely the same path, operators will have a much higher confidence in the traversability of the terrain as the \ac{LUV} will have recently driven in that exact location. 
This allows for higher speeds and may allow for some hazard avoidance systems to operate at lower fidelity saving on computation. 
As the landing sites and habitat will be separated by long distances but occur in consistent locations, it will be worthwhile to expedite rover motion after the first teach/mapping run is completed.

\begin{figure*}[t]
	\centering
	\includegraphics[width=0.95\linewidth]{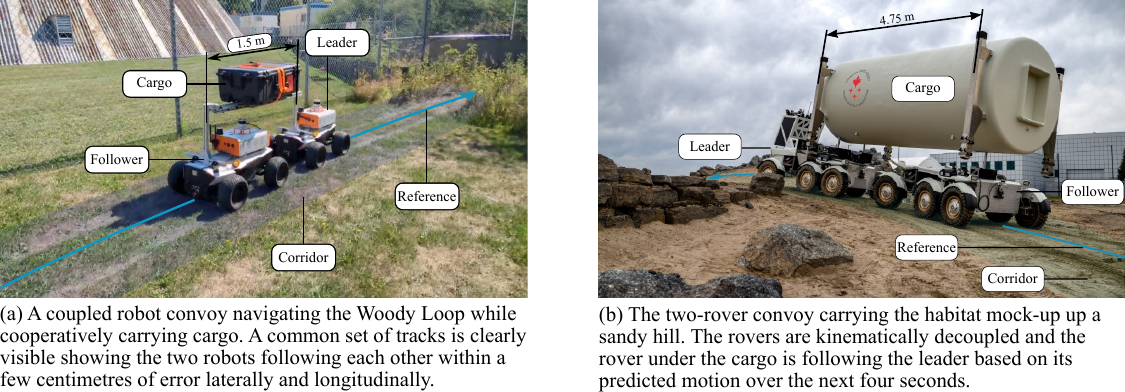}
  \caption{This field report follows the development of a Distributed Model-Predictive Controller and its evaluation on small Ackermann vehicles shown on the left, and scaled up to path-to-spaceflight rovers on the right. The reference path and safe corridor is highlighted on both images and the length of the cargo is provided for relative scale.}
  \label{fig:robots}
\end{figure*}

The cargo's mass and volume makes constructing and launching such a large vehicle increasingly difficult. 
This work aims to establish feasibility of critical subsystems that would allow two rovers to collaboratively transport these large pieces of cargo precisely on the lunar surface. 
Using two, smaller vehicles is attractive for a few reasons. 
First, two rovers can easily accommodate modules that vary in length.
Second, the mass and volume of each rover can be smaller. 
Although sending two rovers to the Moon will be challenging, the smaller size makes each launch individually more achievable.  
Third, having two vehicles that are capable of both coordinated and individual operations increases the overall mission capacity to allow simultaneous missions or pairing one vehicle to follow the astronaut-driven \ac{LTV} \cite{nasa_2020} while the second collects smaller cargo items. 

One concern with carrying cargo between two independent vehicles is the possibility of large loads transferring between the rovers through the cargo.
Accordingly, we design a custom cargo coupling system that allows two rovers to carry the cargo securely but leaves the rovers independent kinematically.  
Decoupling the rovers has a few advantages: ideally no loads are transferred from one rover to the other through the cargo, the control problem is identical whether or not the cargo is loaded, and there is more tolerance to inter-rover misalignment when loading the cargo. 

\begin{figure}
  \centering
  \includegraphics[width=\linewidth]{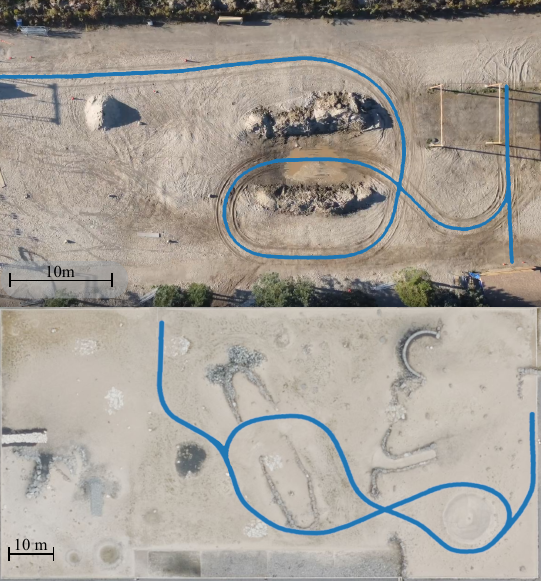}
  \vspace{-5mm}
  \caption{Overhead views of the evaluation sites for the path-to-flight rovers. Scale bars are provided. Top: The sand pit from a drone photo showing the rovers' tracks around the target path. Bottom: A satellite view of the CSA Analogue Terrain with the evaluation route superimposed. }
  \label{fig:eval_paths}
  \vspace{-5mm}
\end{figure}

Recent developments in distributed \acp{MPC} provide an algorithmic baseline that can be expanded to satisfy the requirements of a multi-rover cargo mission \cite{christofides_2013}. 
The field testing highlights coupled direction switches and precise end-of-path alignment that allows the cargo to be docked with a habitat mock-up.
This paper follows the evaluation sequence of developing this system for two path-to-flight rover prototypes.
First, we compare the proposed approach against baselines on small vehicles at the \ac{UTIAS} campus (\autoref{fig:robots}(a)).
Second, we implement and tune the distributed \ac{MPC} on the path-to-flight vehicles to carry a 475 kg piece of cargo in a lunar analogue mission (\autoref{fig:robots}(b)).  
We perform field evaluation at two sites shown in \autoref{fig:eval_paths}: a sand pit, modified to create three-dimensional terrains and the Canadian Space Agency's Analogue Terrain.
The sand pit is 20 m $\times$ 80 m.
The Analogue Terrain is 60 m $\times$ 120 m.
The field testing demonstrated a reliable system that maintained comparable path-tracking errors with or without cargo and the inter-robot Euclidean distance was maintained within the mechanical limits of the coupling with a significant margin. 

\section{RELATED WORK}
\label{sec:related_work}
A convoy refers to a group of vehicles or robots that move in a coordinated manner \cite{nahavandi_2022}.
There is a significant amount of interest in convoying for both trucking (where they are also called platoons \cite{lesch_2022,feng_2024a}) and military vehicles \cite{nahavandi_2022}. 

\citet{zhao_2017} present a convoy of two or three Jeep-style vehicles that drive on gravel roads at approximately 30 km/h with maximum inter-robot distance errors of less than 10 m.
They use a combination of shared GNSS positions fused with camera and lidar tracking of the leader.
Their approach includes the ability of the follower to deviate from the leader's path to avoid encroaching obstacles. 
\citet{vasseur_2004} also follow using cameras to measure the preceding vehicles' positions.
However, for long inter-vehicle distances, the lead vehicle will not be detectable by its followers. 
In this case, the leader sends data about its motion to the followers so that they can localize in a common map to estimate the spacing.
For large separations, path tracking is the only objective.
\citet{bienemann_2025} develop a convoy using two cars driving on gravel roads.
They use a combination of inter-robot localization and direct follower-to-leader measurements. 
While operating only using maps created with simultaneous localization and mapping (SLAM) for localization, they observe that the performance degrades by about 5 cm RMSE compared with direct measurements of the leader.

Our system is designed specifically for teach-and-repeat navigation \cite{furgale_2010}.
Teach and repeat operates in a two-step paradigm where an operator teaches a route by manually driving the robot through the places where it is allowed to navigate. 
While driving, the robot is mapping the environment and recording odometry to define the local shape of the path.
During a repeat, the robot localizes in the map created during the teach and controls the vehicle to track the same path that was taught. 
Teach and repeat uses a topometric pose-graph map that connects places of interest, such as the habitat or landing site, topologically with arbitrary branches similar to the topological map in \autoref{fig:path-network}.
However, each pose is associated with a metric local map \cite{krawciw_2025a} allowing metric localization to be performed in that location. 
In previous testing, robots have operated in \ac{GNSS}-denied environments using stereo cameras \cite{furgale_2010}, lidar \cite{krusi_2017}, and radar \cite{qiao_2025a}. 
In unstructured environments, these robots maintain path-tracking errors on the order of 10 cm or less. 
In teach and repeat, there is a positive feedback of accurate path-tracking: driving in the same location improves localization accuracy and means that the terrain is known to be traversable due to the successful navigation while teaching. 
Teach and repeat is compatible with any type of path-tracking controller, but recent implementations have relied on \ac{MPC} \cite{qiao_2025a,krawciw_2026}.

\ac{MPC} has a rich history in control of non linear systems \cite{christofides_2013}.
Using the Taxonomy of \citet{negenborn_2014}, we are comparing four \acp{MPC}: a Centralized MPC, two \acp{D-MPC}, and a Decentralized \ac{MPC}.
Leader-follower approaches designate one system in the group to be privileged among the rest. 
In this sense, leader-follower approaches are hierarchical, as the leader is making planning decisions based on the desired global state of the system. 
\citet{chen_2010} use a receding-horizon leader-follower approach to control a team of three robots with arbitrary SE(2) offsets from the leader's path in a stable manner. 
They rely on sharing the state history of the lead vehicle.
In a formulation that requires less coordination between vehicles, \citet{gu_2009} develop a flocking controller in which follower vehicles do not know which vehicle in a group is the leader. 
Each vehicle uses the averaged state information from nearby vehicles to estimate what the group's motion is and tries to follow it. 
This provides flexibility and reduces communication requirements with many vehicles. 
While most \ac{D-MPC} methods directly optimize their state based on a shared global state, for systems coupled only by constraints, distributed Tube MPC approaches \cite{trodden_2014} can minimize communication.
Each system selects a stable tube of states in which it will operate. 
Communication is reduced as each system must only share when it has changed tubes.
Recently, formation control for groups of Ackermann vehicles has been approached using deep reinforcement learning instead of \ac{MPC} \cite{zhong_2025}.
Similar to \ac{D-MPC} approaches, \citet{zhong_2025} find that training single-robot path-tracking policies first has the largest impact on shared behaviour in the formation. 

This work investigates a specific application of convoying: cooperative cargo transport. 
For cooperative cargo transport, each robot is coupled in the \ac{MPC} formulation in both the objective function, to minimize the deviation of the cargo separation, and the constraints, as the robots cannot collide or reach mechanical limits of the cargo couplings. 
The cost of deviation is much higher when there is shared cargo than when two vehicles are following each other freely as violating constraints can lead to physical damage to the robots or the cargo.
Depending on how the cargo is constructed, there will be limits on the loads and kinematic degrees of freedom.
For cargo operations, the interaction between the cargo and the robots is an important classifying feature.
Using the taxonomy provided by \citet{tuci_2018}, our cargo transport is \textit{grasped}, i.e., the cargo is kinematically fully defined by the state of the convoy.
For this reason, we omit related works in which robots interact by surrounding or pushing their cargo. 

Breaking down existing approaches to grasped cargo transport, we identify two distinct types: those that allow inter-robot communication and those that do not.
This distinction is important as moving systems face unique challenges such as network fragmentation, high-velocity signal degradation, and outdoor environmental interference \cite{hussein_2022}. 

Systems that do not rely on communication eliminate this bottleneck and rely on reactive control schemes based on inter-robot measurements of distance or transferred force.
In GEOMOVE, \citet{rizzo_2020} develop a controller for large holonomic robots in a factory. 
The follower robot looks for lidar reflectors on the leader to estimate their relative pose. 
They demonstrate that the vehicles can maintain inter-robot separation errors of less than 5 cm. 
These robots use a common 2D lidar map for localization. 
Similarly, \citet{koung_2021} develop a hierarchical controller that uses quadratic programming to solve for joint constraints of holonomic vehicles carrying shared loads. 
The coordinated problem is solved first and then each robot follows its target plan but without inter-robot coordination during navigation. 
This removes the requirement for inter-robot communication but makes the system non-reactive to dynamic effects. 
\citet{xie_2025, rauniyar_2021} use a combination of force sensing and lidar retro-reflector detections to carry a piece of cargo coupling two vehicles. 
The experiments are performed on a smooth indoor surface with holonomic vehicles at low speeds.
Similarly, \citet{huzaefa_2023} use force estimation to control UGVs. 
Their testing is also limited to an indoor location with external state estimates, but they develop a method to control $N$ UGVs that all connect directly to the common cargo.
The known structure of the environment allows for reactive control schemes to dampen small disturbances.

Systems that rely on networking do so primarily to share details of the global state that may be difficult to observe directly from other vehicles or to share predictions \cite{zhang_2019a} that can improve system performance. 
\citet{stroupe_2005} demonstrate the ability to construct small structures using a pair of mobile robots with manipulators. 
A central controller gathers position and force-torque data from both vehicles to generate simultaneous commands. 
Interactions between the cargo and the environment mean that the forces on each vehicle are different making reactive control difficult.
\citet{liu_2023} propose using tracked vehicles with a compliant joint for connecting to cargo in different shapes.
They make use of communication to follow a bi-level controller where the high level considers a vehicle with many steered wheels, and then each robot follows the reference angle and velocity to mimic a wheel. 
The approach allows for existing non-holonomic controllers to be used at the high level and separates some complexity for inter-robot communication.
\citet{zhang_2025} construct a pair of Ackermann robots and employ independent sliding-mode control on each vehicle to track a setpoint in front of the leader and the desired following position relative to that leader for the follower. 
Wireless communication between the two vehicles allows for the leader's position to be sent to the follower. 
The flexibility of task specification and robot arrangement makes wireless communication attractive for this cargo transport case despite the challenges of maintaining a reliable link. 

Space-focused cargo transportation has been less explored as most missions focus on exploration. 
\citet{huntsberger_2004} develop a pair of vehicles that carry long beams together.
With a view towards carrying materials that would be required to construct buildings on future Martian mission, the two vehicles contain symmetric couplings that provide rotational independence and one prismatic degree of freedom.
Interestingly, due to this symmetric coupling, their cargo is under-constrained and is free to slide along the axis joining the robots.
This adds an adjustment step, where periodically the vehicles have to move along that axis to maintain the desired position and load distribution between both rovers.
This coupling approach differs from our own to create a homogeneous pair vehicles, but at the expense of transport precision.
The ATHLETE robot \cite{wilcox_2009} features six wheel-on-limbs with a central hexagonal platform for carrying cargo.
On Earth, ATHLETE could carry up to 300 kg of payload \cite{wilcox_2011}.
Large cargo cylinders mocking habitat modules were docked together to create a habitat structure in a large-scale field test. 
The wheel-on-limb design enables ATHLETE to move in redundant ways.
Under normal operation, the platform rolls on its six wheels and steers holonomically by twisting the arm.
The arms allow for clearing much larger obstacles than the wheels alone.
This minimizes robot mass because smaller wheels and motors can be used.
An extension of ATHLETE was proposed to collect and deliver payloads.
Tri-ATHLETE splits an ATHLETE robot into two half robots with three limbs each. 
The two vehicles approach a cargo from either side and latch onto it.
The pallet of the cargo forms the central platform and rigidly links the two Tri-ATHLETEs into one functioning system. 
Although the half-rovers are less stable, they can still safely navigate without payloads.
Once the two robots are rigidly coupled, they are controlled as a single system. 
This design requires that the cargo forms a structural member of the final system which may provide less flexibility as cargo requirements evolve. 
The relatively flat terrain of the Artemis landing sites \cite{pena-asensio_2025} means that the extra complexity of the drive system should not be required. 
Simpler vehicles with large payload capacity and operational redundancy will be well suited for multi-purpose long-duration missions. 
As inter-robot communication will be required when there is no cargo that connects the rovers, the flexibility of a D-MPC that does not explicitly capture the cargo in its formulation is well suited for this problem.

\section{METHOD}
\label{sec:method}
Throughout this paper, we consider a robot convoy consisting of several Ackermann-steered vehicles.
For the sake of conciseness, all methods are presented with a convoy of two robots, consisting of one leader and one follower as depicted in \autoref{fig:robots}.
We discuss the methods' capabilities to scale to multiple robots at the end of this section.

\begin{figure*}[t]
	\centering
	\includegraphics[width=1\linewidth]{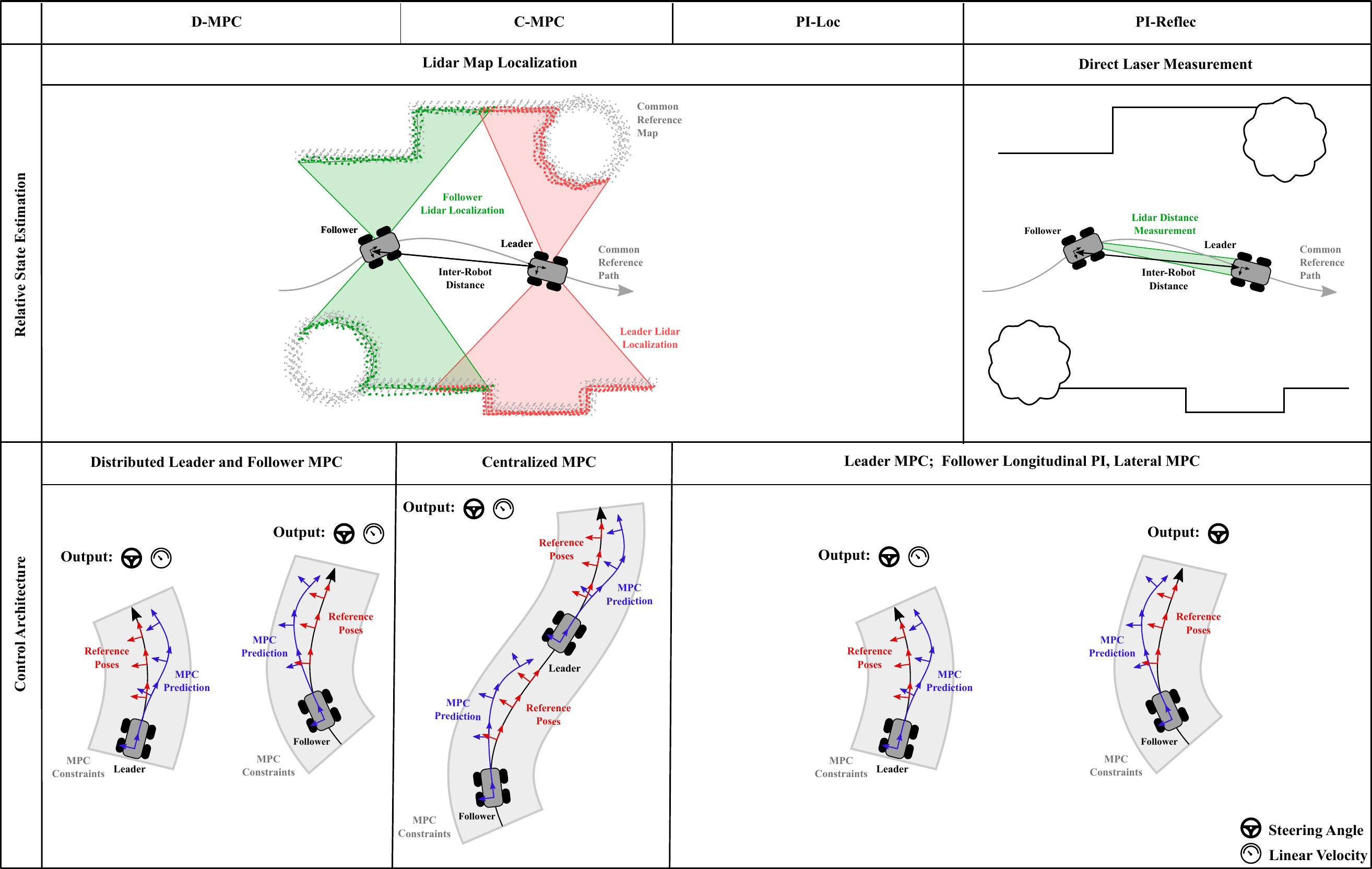}
	\vspace{-5mm}
	\caption{The four controllers share similarities but are divided by how the inter-robot distance is measured and what variables are free for the MPC to optimize. All methods allow the steering angle to be controlled (steering wheel icon) but only some allow the linear velocity (gauge icon) to vary.}
	\label{fig:methods_overview}
\end{figure*}

The current state of each robot, $\mathbf{T}(t) \in \text{SE}(2)$, is the pose at the center of the rear wheels with position $\mathbf{r}(t) =  \left[ x(t) \quad y(t) \right]^T  \in \mathds{R}^2$ and orientation $\theta(t) \in \mathds{R}$.
The control inputs for each robot include its forward velocity and steering angle 
$\mathbf{u}(t) = \left[v(t) \quad \delta(t)\right]^T \in \mathds{R}^2$.
It is assumed that the motion of each robot can be described using a standard kinematic non-slip bicycle model \cite{kong_2015} with the ordinary differential equation (ODE),
\begin{align}
	f(\mathbf{T}(t),\mathbf{u}(t)) = \begin{bmatrix}
		\dot{x}(t) \\ \dot{y}(t) \\ \dot{\theta}(t)
	\end{bmatrix} = \begin{bmatrix}
		v(t)\cdot\mathrm{sin}(\theta(t)) \\ v(t)\cdot\mathrm{cos}(\theta(t)) \\ \frac{v(t)}{L} \cdot \mathrm{tan}(\delta(t))
	\end{bmatrix},
\end{align}
where $L$ is the separation between front and rear wheels.
The subscripts $\ell$ and $f$ will be used to denote which states and inputs correspond to the leader and follower, respectively.

Our objective is to determine suitable control inputs for both robot systems, such that they accurately track a predefined reference path (in the following also called teach path) while simultaneously maintaining a desired inter-robot distance.
The distance can be defined either in Euclidean space or along the arclength of the teach path.
The former is applicable to the primary use case in which the robots are physically coupled to jointly transport cargo.
In this setting, joints that constrain position but allow free rotation reduce the control problem to keeping the follower at the desired Euclidean distance from the leader.

\subsection{System Overview}

For implementation, we use a model-predictive control scheme with a control horizon of $K$ robot states  $\mathbf{T}_k = \mathbf{T}(t_k)$, where the sampling times $t_k$ are spaced by $\Delta T$.
The MPC finds the optimal sequence of control inputs $\mathbf{u}_k = \mathbf{u}(t_k)$ that minimize a cost function for path and inter-robot distance tracking, subject to constraints.
This is guided by defining a set of suitable, discrete reference poses $\mathbf{T}_{\mathrm{ref},k}$ for the robots along the teach path.
A safe corridor, $\mathcal{X}_\text{corr}$, bounds the teach path, defining the maximum lateral error along the path.
In this work, we focus on developing a distributed system in which each robot solves its own localization and control problems, requiring minimal communication between individual vehicles.
Information flows only from the leader to the follower, enabling distribution across many robots.
Additional MPC configurations are considered as baseline comparisons.
\autoref{fig:methods_overview} provides an overview of the different control architectures.
The remainder of this section details the individual components of the proposed system.

\subsection{Localization}

Each robot runs the Lidar Teach \& Repeat (LT\&R) navigation stack \cite{furgale_2010}.
The localization is performed in $\text{SE}(3)$; however, we project onto $\text{SE}(2)$ for planning and control purposes.
We assume that the teach phase has already been completed (using any one robot) and that each robot in the convoy maintains a local copy of the resulting teach path and submaps.
Using this data, each robot is able to localize itself with respect to the teach path individually.
While we use LiDAR localization for this work, the control approach is agnostic to the particular sensor modality.
For details on teach path and submap construction and the localization capabilities of LT\&R, we refer readers to \cite{wu_2022}.

\subsection{Reference Pose Planning}

Using the successful localization of each robot, $K$ discrete reference poses $\mathbf{T}_{\mathrm{ref},k}$ at times $t_k$  for both leader and follower are defined along the teach path in $\text{SE}(3)$.

\subsubsection{Leader Reference Poses}

The leader reference poses are defined using a desired forward velocity of the robot convoy.
Given the desired convoy speed and the leader’s current state (extrapolated from the most recent localization under a constant-velocity assumption), the reference poses can be defined along the teach path in a straightforward manner.
We note that this reference pose selection is the baseline approach implemented in the latest version of LT\&R \cite{qiao_2025a}.

\subsubsection{Follower Reference Poses}

For the follower reference pose planning, two alternative methods are considered.
Both take the desired inter-robot distance explicitly into account.

\paragraph{Distance-Aware Pose Selection}

The first method relies on a communication link between leader and follower, allowing the follower to compute reference poses that maintain the desired inter-robot distance using the latest information about the leader.
The simplest approach uses the planned leader reference poses.
For improved performance, the leader's state estimate and anticipated motion, obtained from its most recent MPC solution, can be used instead.
Interpolation of the MPC rollout provides the anticipated leader state at the follower's reference times.
Once the leader states at the desired times are determined, the follower reference poses are defined along the teach path such that they satisfy the inter-robot distance, either in Euclidean space or along the path's arclength.
Because this method depends on the quality of the leader's state estimates and MPC tracking, errors can arise if these are inaccurate or biased.

\paragraph{PI Feedback Controller Pose Selection}
To evaluate this effect, a baseline reference pose selection method implements a proportional-integral (PI) feedback controller to control directly the follower's speed based on the inter-robot distance error:
\begin{align}
	v_\mathrm{ref,f}(t) &=  K_p e_\mathrm{dist}(t) + K_i \int_0^t e_\mathrm{dist}(\tau)d\tau,
\end{align}
where $e_\mathrm{dist}$ denotes the error from the desired inter-robot distance.
The velocity obtained from this controller is assumed constant across the follower’s reference poses, making the reference pose selection analogous to that of the leader.
The main advantage of this approach is that it explicitly incorporates the inter-robot distance, allowing direct corrective action through proportional and integral feedback.

\subsection{Model-Predictive Control Architectures}
Three nonlinear MPC architectures are considered, each formulated in $\text{SE}(2)$ by projecting the robot's reference poses into the plane tangent to the nearest path segment.
Two of those architectures are distributed, while an additional centralized method is investigated for comparison.
Each architecture considers a common set of cost terms and constraints.
The cost terms are given as
\begin{subequations}
\begin{align}
	J_{\mathrm{ref},k} &= {\log\left(\mathbf{T}_{\mathrm{ref},k}\mathbf{T}_{k}^{-1}\right)^\vee}^T\mathbf{Q}_\mathrm{ref}^{-1}\log\left(\mathbf{T}_{\mathrm{ref},k}\mathbf{T}_{k}^{-1}\right)^\vee, \\
	J_{\mathrm{cont},k} &= \mathbf{u}_{k}^T \mathbf{Q}_\mathrm{cont}^{-1}\mathbf{u}_{k}, \\
	J_{\mathrm{acc},k} &= \left(\mathbf{u}_{k} - \mathbf{u}_{{k-1}}\right)^T \mathbf{Q}_\mathrm{acc}^{-1}\left(\mathbf{u}_{k} - \mathbf{u}_{{k-1}}\right), \\
	J_{\mathrm{dist},k} &= ||\mathbf{r}_{\mathrm{l},k} - \mathbf{r}_{\mathrm{f},k}||_2^T \mathbf{Q}_\mathrm{dist}^{-1}||\mathbf{r}_{\mathrm{l},k} - \mathbf{r}_{\mathrm{f},k}||_2,
\end{align}
\end{subequations}
where $J_\mathrm{ref}$ controls reference pose tracking, $J_\mathrm{cont}$ reduces control-input magnitude, $J_\mathrm{acc}$ reduces acceleration, and $J_\mathrm{dist}$ controls the inter-robot distance.
$\log\left(\mathbf{T}\right)^\vee$ transforms an element $\mathbf{T} \in \text{SE}(2)$ into the $3\times1$ element of the Lie algebra $\mathfrak{se}(2)$.
The weighting matrices $\mathbf{Q}$ for each cost term are constant.

The constraints are given as
\begin{subequations}
\begin{align}
	g_{\mathrm{model},k}:& \quad \mathbf{T}_k - \int_{t_{k-1}}^{t_k}f(\mathbf{T}(t),\mathbf{u}(t)) dt = \mathbf{0}, \label{eq:motion_constraint}\\
	g_{\mathrm{cont},k}:& \quad  \mathbf{u}_\mathrm{min} < \mathbf{u}_k < \mathbf{u}_\mathrm{max}, \\
	g_{\mathrm{acc},k}:&  \quad \Delta\mathbf{u}_\mathrm{min} < \mathbf{u}_k - \mathbf{u}_{k-1} < \Delta\mathbf{u}_\mathrm{max}, \\
	g_{\mathrm{corr},k}:&  \quad \mathbf{r}_k \in \mathcal{X}_\mathrm{corr}, \\
	g_{\mathrm{dist},k}:&  \quad d_\mathrm{min} <  ||\mathbf{r}_{\mathrm{l},k} - \mathbf{r}_{\mathrm{f},k}||_2 < d_\mathrm{max},
\end{align}
\end{subequations}
where $g_{\mathrm{model}}$ enforces adherence to the kinematic motion model, $g_\mathrm{cont}$ limits the control-input magnitude, and $g_\mathrm{acc}$ limits acceleration.
Additionally, $g_\mathrm{corr}$ ensures that the robots remain within the safe corridor $\mathcal{X}_\mathrm{corr}$ around the path and $g_\mathrm{dist}$ maintains a collision-free inter-robot distance, given minimum and maximum distances $d_\mathrm{min}$ and $d_\mathrm{max}$.
During implementation, \eqref{eq:motion_constraint} is solved numerically, using a Runge-Kutta (4,5) method, assuming constant control inputs over interval $\left[t_{k-1},t_k\right]$ and accounting for a first-order lag.

Next, we discuss how these cost terms and constraints are assembled into optimization problems for each controller.
The resulting optimizations are implemented using CasADi \cite{andersson_2019}.

\subsubsection{Distributed MPC}

We propose a distributed MPC architecture, which solves for the control inputs of the leader and follower individually.
This means that each robot solves their respective control optimization problem locally.
The optimization problem for the leader can be written as
\begin{align}
	\label{eq:leader_mpc}
	\argmin_{\mathbf{T}_{\mathrm{l},k},\mathbf{u}_{\mathrm{l},k}} \quad &\sum_{k=1}^K  J_{\mathrm{ref,l},k} + J_{\mathrm{cont,l},k} + J_{\mathrm{acc,l},k} \\
	& \mathrm{s.t.} \quad \forall k \in 1,2,...,K \notag\\
	& g_{\mathrm{model,l},k},~ g_{\mathrm{cont,l},k},~ g_{\mathrm{acc,l},k},~ g_{\mathrm{corr,l},k}, \notag
\end{align}
which optimizes for the leader's states, $\mathbf{T}_{\mathrm{l},k}$, and inputs, $\mathbf{u}_{\mathrm{l},k}$, at each discrete time $t_k$ during the prediction horizon.
The objective function includes the reference pose tracking as well as control-input magnitude, and acceleration for the leader.
The motion model, control-input magnitude, acceleration, and corridor constraints are active.

Analogously, the optimization problem for the follower can be written as
\begin{align}
	\argmin_{\mathbf{T}_{\mathrm{f},k},\mathbf{u}_{\mathrm{f},k}} \quad&\sum_{k=1}^K  J_{\mathrm{ref,f},k} + J_{\mathrm{cont,f},k} + J_{\mathrm{acc,f},k} + J_{\mathrm{dist},k} \\
	& \mathrm{s.t.} \quad \forall k \in 1,2,...,K \notag\\
	~&g_{\mathrm{model,f},k}, g_{\mathrm{cont,f},k},~ g_{\mathrm{acc,f},k},~ g_{\mathrm{corr,f},k},~g_{\mathrm{dist},k}. \notag
\end{align}
The objective function includes the additional cost term and constraint for the inter-robot distance, $J_{\mathrm{dist},k}$.
The key innovation in this paper is passing the MPC solution from the leader back to the follower. 
Interpolating the real solution has better performance than the reference poses or the current state alone.
The computation of these terms as well as the definition of the reference poses utilize the last reported localization result and MPC prediction from the leader.
This method requires a communication link.
In the discussion, this method is abbreviated D-MPC.

\begin{figure*}[t]
	\centering
    \includegraphics[width=0.8\linewidth]{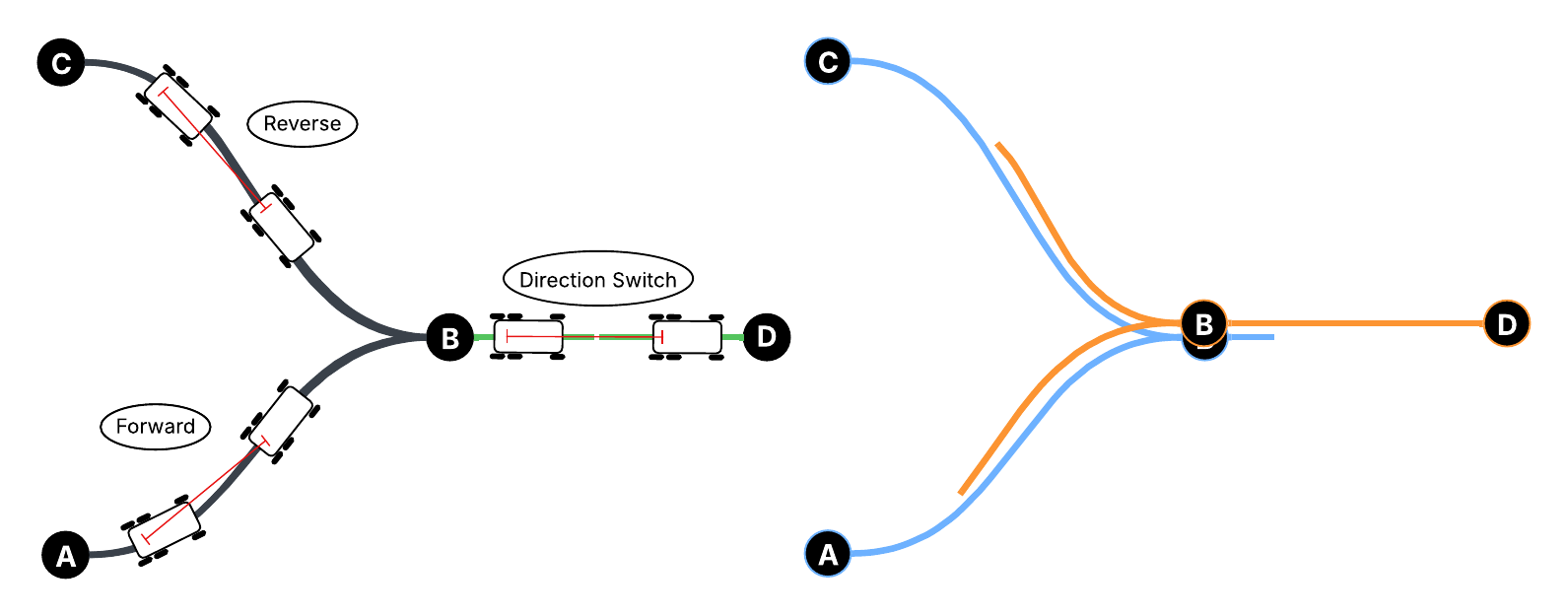}
    \caption{Two coupled rovers passing through a merge-point direction switch. A single robot could travel directly from A to B to C. When the rovers are coupled by cargo, the direction switch requires an extra segment extending from B. The segment from B to D (\textit{dashed green line}) must be longer than the maximum inter-robot distance. On the right, the leader's path (\textit{solid orange}) is different than the follower's (\textit{solid blue}). The short path past B for the follower occurs because the length of segment B-D is based on the maximum inter-robot distance not the nominal target. }
    \label{fig:convoy_switch}
\end{figure*}

\subsubsection{Centralized MPC}
As a non-scalable baseline to the proposed distributed MPC architecture, a centralized implementation is also considered.
Here, the states and control inputs to both robots are solved in one optimization problem, formulated as
\begin{align}
	\argmin_{\substack{\mathbf{T}_{\mathrm{l},k},\mathbf{u}_{\mathrm{l},k}\\\mathbf{T}_{\mathrm{f},k},\mathbf{u}_{\mathrm{f},k}}} \quad &\sum_{k=1}^K  J_{\mathrm{ref,l},k} + J_{\mathrm{cont,l},k}  + J_{\mathrm{acc,l},k} \\
	&\quad + J_{\mathrm{ref,f},k} + J_{\mathrm{cont,f},k} + J_{\mathrm{acc,f},k} + J_{\mathrm{dist},k} \notag\\
	& \mathrm{s.t.} \quad \forall k \in 1,2,...,K \notag\\
	~ &g_{\mathrm{model,l},k},g_{\mathrm{cont,l},k},~ g_{\mathrm{acc,l},k},~ g_{\mathrm{corr,l},k}, \notag\\
	~ &g_{\mathrm{model,f},k},g_{\mathrm{cont,f},k},~ g_{\mathrm{acc,f},k},~ g_{\mathrm{corr,f},k},~ g_{\mathrm{dist},k}. \notag
\end{align}
This problem includes all optimization variables, cost terms and constraints from the two distributed ones.
The distance cost term and constraints are adapted to use the distance between the variable leader and follower poses.
Additionally, the follower's reference poses are defined with respect to the leader's reference poses, because the leader's MPC has not yet been solved.
The solution can be computed on either the leader or follower and requires a communication link.
In the discussion, this method is abbreviated as C-MPC.

\subsubsection{Independent Lateral and Longitudinal Control}

Lastly, we consider a control architecture that utilizes the reference pose selection based on the PI feedback controller.
This architecture is distributed, and the optimization problem for the leader remains \eqref{eq:leader_mpc}.
The reference poses for the follower are defined according to the velocity computed by the PI controller on the inter-robot distance.
Additionally, this velocity is added as an equality constraint and the optimization problem for the follower MPC becomes
\begin{align}
	\argmin_{\mathbf{T}_{\mathrm{f},k},\mathbf{u}_{\mathrm{f},k}} \quad&\sum_{k=1}^K J_{\mathrm{ref,f},k} + J_{\mathrm{cont,f},k} + J_{\mathrm{acc,f},k} \\
	& \mathrm{s.t.} \quad \forall k \in 1,2,...,K \notag\\
	~ &g_{\mathrm{model,f},k},g_{\mathrm{cont,f},k},~ g_{\mathrm{acc,f},k},~ g_{\mathrm{corr,f},k},~v_{\text{f},k} = v_\text{ref,f}. \notag
\end{align}
Due to the linear velocity constraint, the MPC can select only the steering angles.
This decouples the lateral and longitudinal control of the robot, with the former being selected by the MPC and the latter by the PI controller.

We evaluate two cases for this controller.
In the first, the inter-robot distance is computed from the most recent localization results of the leader and follower, requiring a communication link.
In the discussion, this is abbreviated PI-Loc.
In the second, the distance is measured directly, using a laser rangefinder, eliminating the need for communication.
In the discussion, this is abbreviated PI-Reflec.

\subsection{Scaling to $N$ Robots}

The primary advantage of distributed control approaches is scalability. 
Adding another follower to the convoy is straightforward, as information flows only from the leader to the follower.
Technically, each follower can consider any existing robot in the convoy as their leader.
In practice, the most stable configuration uses one common leader for every follower, so disturbances in one follower's motion will not propagate through the chain of robots.
However, depending on the convoy length and chosen approach, communication links or distance measurements could make a chained following pattern more desirable.

The C-MPC approach is difficult to scale. 
Adding more coupled variables to the optimization problem increases complexity and solve time. 
While this effect is insignificant for two robots, quadratic scaling is expected from the utilized \textit{ipopt} solver \cite{wachter_2006}.
This may limit the effectiveness for large convoys. 
Additional practical challenges include the required \textit{a priori} knowledge of the robot kinematics of every vehicle in the convoy and the software engineering needed to scale to an arbitrary number of followers.

\subsection{Direction Switches}
Unlike a single vehicle navigating within a network of paths, a coupled convoy must switch direction in a coordinated manner. 
See \autoref{fig:convoy_switch} for an example of coupled convoy switching directions at a branch point within the taught path. 
At a direction switch, the follower will not travel as far along the path as the leader. 
Once the convoy starts reversing, the role of the leader and follower vehicles must switch. 
These effects constrain the places where direction switches can occur. 
For a desired direction switch location to be valid, the path must continue for longer than the maximum possible separation between the two vehicles. 
During the route-planning stage, the planner considers the current position of all vehicles and checks that proposed direction switches are valid. 
This constraint prevents the convoy from taking some routes that a single rover could traverse.  

We note that for uncoupled convoys there are different patterns to traversing direction switches.
We leave exploration of uncoupled convoy direction switching to future work.

\section{CONTROLLER COMPARISON SETUP}
\label{sec:hunter_experiments}
To compare the controllers, we use a pair of AgileX Hunter 2.0 vehicles, shown in \autoref{fig:robots}(a). 
These robots have Ackermann steering and rear-wheel drive. 
An Ouster OS-0 128-beam lidar is mounted on top of each vehicle and is used for localization and mapping. 
Kinematically, the cargo has a prismatic joint along its length with a travel range of $\pm$27 cm. 
A loose pin joint on each vehicle connects it to the cargo.
The loose pins act as spherical joints; although, the flatness of the cargo means that roll is minimized but not constrained. 
Routers on each robot connect to each other and messages are passed using the ROS2 Zenoh \cite{_i} middleware.
A retro-reflective panel is placed on the back of the leader.
Segmenting these points and using a robust plane-fit provides the direct measurement of the leader's distance from the follower.
This signal is used to evaluate the quality of all controllers.
This measurement was validated by a high-framerate video of a tape measure as it extends while the vehicles were in motion.

Four loops are evaluated three times for each controller. 
The convoy has nominal forward speed of 0.7 m/s.
The terrain has a significant impact on the quality of single-robot navigation so we choose paths that are progressively more difficult for the system to handle. 
In the Woody loop, the follower sometimes loses track of the leader's retro-reflective panel.
As this distance is our evaluation signal, the Woody Loop was used only for qualitative analysis.
A summary of the paths is provided in Table \ref{tab:loops}.
The supplementary video provides samples of the convoy operating in all environments. 

\newcommand{\loopTableHeader}{
	\scriptsize
	\begin{tabularx}{\linewidth}{q{0.3cm}|p{1.6cm}|q{1.5cm}|q{0.8cm}|q{0.7cm}|q{1.0cm}|}
		\cline{2-6}
		& \textbf{Relative \linebreak State Est.}
		& \multicolumn{3}{c|}{ Lidar Map Localization} 
		& \multicolumn{1}{p{1.0cm}|}{\centering Laser Meas.} \\ \cline{2-6}
		& \textbf{Controller} & D-MPC (Ours) & C-MPC & PI-Loc & PI-Reflec \\ \clineB{2-6}{3}
	\end{tabularx}
	\normalsize
	\vspace{2pt}
}


\newcommand{\loopTable}[1]{%
	\scriptsize
	\begin{tabularx}{\linewidth}{p{0.3cm}|p{1.6cm}|p{1.0cm}|p{1.0cm}|p{0.8cm}|p{1cm}|}
		\multicolumn{6}{c}{\textbf{#1}} \\ \clineB{1-6}{3}
		\multicolumn{1}{|p{0.3cm}|}{\multirow{4}{*}{ \rotatebox[origin=r]{90}{\parbox[l]{12mm}{\centering \vspace{-0.2cm} \textbf{Euclidean \linebreak Distance }}}}} &
		Mean Err. & & & &  \\ \cline{2-6}
		\multicolumn{1}{|p{0.3cm}|}{}&RMSE & & & & \\ \cline{2-6}
		\multicolumn{1}{|p{0.3cm}|}{}&Std. Dev. & & & & \\ \cline{2-6}
		\multicolumn{1}{|p{0.3cm}|}{}&Max Err. & & & & \\ \clineB{1-6}{3}
		 \multicolumn{1}{|p{0.3cm}|}{\multirow{4}{*}{ \rotatebox[origin=r]{90}{\parbox[l]{12mm}{\centering \vspace{-0.2cm} \textbf{Path \linebreak Tracking }}}}}& Lead. RMSE & & & & \\ \cline{2-6}
		\multicolumn{1}{|p{0.3cm}|}{}&Lead. Max Err. & & & & \\ \cline{2-6}
		\multicolumn{1}{|p{0.3cm}|}{}&Fol. RMSE & & & & \\ \cline{2-6}
		\multicolumn{1}{|p{0.3cm}|}{}&Fol. Max Err. & & & & \\ \clineB{1-6}{3}
	\end{tabularx}
	\normalsize
   \vspace{1pt}
}

\newcommand{\loopTableParking}{%
	\scriptsize
	\begin{tabularx}{\linewidth}{p{0.3cm}|p{1.6cm}|q{1.5cm}|q{0.8cm}|q{0.7cm}|q{1cm}|}
		\multicolumn{6}{c}{\textbf{Parking Loop}} \\ \clineB{1-6}{3}
		\multicolumn{1}{|p{0.3cm}|}{\multirow{4}{*}{ \rotatebox[origin=r]{90}{\parbox[l]{12mm}{\centering \vspace{-0.2cm} \textbf{Euclidean \linebreak Distance }}}}} &
		Mean Err. & -2.1 & -4.7 & 0.7 & \textbf{0.4} \\ \cline{2-6}
		\multicolumn{1}{|p{0.3cm}|}{}&RMSE & 3.9 & 6.0 & 6.9 & \textbf{3.2}\\ \cline{2-6}
		\multicolumn{1}{|p{0.3cm}|}{}&Std. Dev. & \textbf{3.2} & 3.7 & 6.9 & \textbf{3.2}\\ \cline{2-6}
		\multicolumn{1}{|p{0.3cm}|}{}&Max Err. & \textbf{12.9} & 18.7 & 27.6 & 19.4\\ \clineB{1-6}{3}
		 \multicolumn{1}{|p{0.3cm}|}{\multirow{4}{*}{ \rotatebox[origin=r]{90}{\parbox[l]{12mm}{\centering \vspace{-0.2cm} \textbf{Path \linebreak Tracking }}}}}& Lead. RMSE & 2.9 & \textbf{2.7} & \textbf{2.7} & \textbf{2.7} \\ \cline{2-6}
		\multicolumn{1}{|p{0.3cm}|}{}&Lead. Max Err. & 7.4 & 7.4 & 7.2 & \textbf{7.1} \\ \cline{2-6}
		\multicolumn{1}{|p{0.3cm}|}{}&Fol. RMSE & 5.3 & 5.3 & 5.3 & \textbf{5.2} \\ \cline{2-6}
		\multicolumn{1}{|p{0.3cm}|}{}&Fol. Max Err. & 10.2 & 10.7 & \textbf{8.7} & 10.3 \\ \clineB{1-6}{3}
	\end{tabularx}
	\normalsize
   \vspace{1pt}
}

\newcommand{\loopTableMars}{%
	\scriptsize
	\begin{tabularx}{\linewidth}{p{0.3cm}|p{1.6cm}|q{1.5cm}|q{0.8cm}|q{0.7cm}|q{1cm}|}
		\multicolumn{6}{c}{\textbf{Dome Loop}} \\ \clineB{1-6}{3}
		\multicolumn{1}{|p{0.3cm}|}{\multirow{4}{*}{ \rotatebox[origin=r]{90}{\parbox[l]{12mm}{\centering \vspace{-0.2cm} \textbf{Euclidean \linebreak Distance }}}}} &
		Mean Err. &  -2.6 & -2.6 & 1.8 & \textbf{0.3} \\ \cline{2-6}
		\multicolumn{1}{|p{0.3cm}|}{}&RMSE & 5.0 & \textbf{4.9} & 6.1 & 5.0 \\ \cline{2-6}
		\multicolumn{1}{|p{0.3cm}|}{}&Std. Dev. & 4.3 & 4.1 & 5.8 & \textbf{3.4}\\ \cline{2-6}
		\multicolumn{1}{|p{0.3cm}|}{}&Max Err. & -17.3 & \textbf{-16.5} & 26.6 & 18.8\\ \clineB{1-6}{3}
		 \multicolumn{1}{|p{0.3cm}|}{\multirow{4}{*}{ \rotatebox[origin=r]{90}{\parbox[l]{12mm}{\centering \vspace{-0.2cm} \textbf{Path \linebreak Tracking }}}}}& Lead. RMSE & 2.8 & \textbf{2.7} & 2.9 & 2.9 \\ \cline{2-6}
		\multicolumn{1}{|p{0.3cm}|}{}&Lead. Max Err. & 9.8 & \textbf{9.2} & 10.7 & 10.5 \\ \cline{2-6}
		\multicolumn{1}{|p{0.3cm}|}{}&Fol. RMSE  & 4.5 & 4.6 & \textbf{4.4} & \textbf{4.4} \\ \cline{2-6}
		\multicolumn{1}{|p{0.3cm}|}{}&Fol. Max Err. & 10.3 & 10.3 & 10.0 & \textbf{9.8} \\ \clineB{1-6}{3}
	\end{tabularx}
	\normalsize
   \vspace{1pt}
}

\newcommand{\loopTableHillside}{%
	\scriptsize
	\begin{tabularx}{\linewidth}{p{0.3cm}|p{1.6cm}|q{1.5cm}|q{0.8cm}|q{0.7cm}|q{1cm}|}
		\multicolumn{6}{c}{\textbf{Hillside Loop}} \\ \clineB{1-6}{3}
		\multicolumn{1}{|p{0.3cm}|}{\multirow{4}{*}{ \rotatebox[origin=r]{90}{\parbox[l]{12mm}{\centering \vspace{-0.2cm} \textbf{Euclidean \linebreak Distance }}}}} &
		Mean Err.  & -3.5 & -3.3 & 0.7 & \textbf{0.3} \\ \cline{2-6}
		\multicolumn{1}{|p{0.3cm}|}{}&RMSE & 6.6 & 6.6 & 7.5 & \textbf{4.2} \\ \cline{2-6}
		\multicolumn{1}{|p{0.3cm}|}{}&Std. Dev. & 5.6 & 5.8 & 7.5 & \textbf{4.2} \\ \cline{2-6}
		\multicolumn{1}{|p{0.3cm}|}{}&Max Err. & 19.5 & -29.8 & -27.8 & \textbf{18.2}\\ \clineB{1-6}{3}
		 \multicolumn{1}{|p{0.3cm}|}{\multirow{4}{*}{ \rotatebox[origin=r]{90}{\parbox[l]{12mm}{\centering \vspace{-0.2cm} \textbf{Path \linebreak Tracking }}}}}& Lead. RMSE & 7.2 & \textbf{6.7} & 6.9 & 7.1 \\ \cline{2-6}
		\multicolumn{1}{|p{0.3cm}|}{}&Lead. Max Err. & 28.1 & 26.2 & \textbf{22.7} & 26.7 \\ \cline{2-6}
		\multicolumn{1}{|p{0.3cm}|}{}&Fol. RMSE & 6.1 & 6.4 & \textbf{5.9} & 6.1 \\ \cline{2-6}
		\multicolumn{1}{|p{0.3cm}|}{}&Fol. Max Err. & 22.4 & 20.5 & 21.2 & \textbf{18.7} \\ \clineB{1-6}{3}
	\end{tabularx}
	\normalsize
   \vspace{1pt}
}

\newcommand{\loopTableTennis}{%
	\scriptsize
	\begin{tabularx}{\linewidth}{p{0.3cm}|p{1.6cm}|q{1.5cm}|q{0.8cm}|q{0.7cm}|q{1cm}|}
		\multicolumn{6}{c}{\textbf{Tennis Loop}} \\ \clineB{1-6}{3}
		\multicolumn{1}{|p{0.3cm}|}{\multirow{4}{*}{ \rotatebox[origin=r]{90}{\parbox[l]{12mm}{\centering \vspace{-0.2cm} \textbf{Euclidean \linebreak Distance }}}}} &
		Mean Err. & -2.2 & -2.2 & 1.3 & \textbf{0.5} \\ \cline{2-6}
		\multicolumn{1}{|p{0.3cm}|}{}&RMSE & 4.3 & 4.5 & 6.1 & \textbf{3.7} \\ \cline{2-6}
		\multicolumn{1}{|p{0.3cm}|}{}&Std. Dev. & \textbf{3.7} & 3.9 & 5.9 & \textbf{3.7} \\ \cline{2-6}
		\multicolumn{1}{|p{0.3cm}|}{}&Max Err. & \textbf{-14.8} & -15.1 & 26.7 & 19.1\\ \clineB{1-6}{3}
		 \multicolumn{1}{|p{0.3cm}|}{\multirow{4}{*}{ \rotatebox[origin=r]{90}{\parbox[l]{12mm}{\centering \vspace{-0.2cm} \textbf{Path \linebreak Tracking }}}}}& Lead. RMSE & 5.1 & \textbf{4.8} & 5.4 & 5.6 \\ \cline{2-6}
		\multicolumn{1}{|p{0.3cm}|}{}&Lead. Max Err. & 13.7 & \textbf{11.4} & 14.6 & 15.4 \\ \cline{2-6}
		\multicolumn{1}{|p{0.3cm}|}{}&Fol. RMSE & 5.4 & 5.6 & 5.1 & \textbf{4.9} \\ \cline{2-6}
		\multicolumn{1}{|p{0.3cm}|}{}&Fol. Max Err. & 13.2 & 14.7 & 13.1 & \textbf{11.2} \\ \clineB{1-6}{3}
	\end{tabularx}
	\normalsize
   \vspace{1pt}
}

\begin{table}[t]
\caption{Evaluation Route Summaries}
\centering
\scriptsize
\begin{tabularx}{\linewidth}{|c|c|c|o{1.0cm}|C|}
   \hline
   \textbf{Route} & \textbf{Length} & \textbf{\# Trials} & \textbf{Total\linebreak Distance} & \textbf{Terrain}\\
   \hline
   Tennis & 113 m & 12 & 1.7 km & Cracked but smooth cement.\\
   \hline
   Parking & 200 m & 12 & 2.4 km & Asphalt parking lot.\\
   \hline
   Dome & 260 m & 15 & 3.9 km & Mowed grass lawn.\\
   \hline
   Hillside & 180 m & 12 & 2.51 km & Mowed grass. $\pm10^\circ$ roll and pitch inclines.\\
   \hline 
   Woody & 278 m & 4 & 1.11 km & Tall grass, branches, potholes.\\
   \hline
\end{tabularx}
\label{tab:loops}
\vspace{-5mm}
\end{table}

\begin{table*}[t]
   \centering
   \caption{Error metrics for four convoyed routes using the Hunters. Each column combines results from three repetitions on each route. All values are in cm.}
   \noindent
   \begin{tabularx}{\textwidth}{X X}
   	\centering
      \loopTableHeader & \loopTableHeader \\
      \loopTableParking & \loopTableTennis \\ 
      \loopTableMars & \loopTableHillside
   \end{tabularx}
   \label{tab:results}
   \vspace*{-3mm}
\end{table*}

\section{COMPARISON RESULTS}
\label{sec:hunter_quant_results}

The results in Table \ref{tab:results} quantify high-level patterns that were anticipated. 
The Centralized and Distributed MPCs have no integral-type term and exhibit a bias on all loops tested.
The PI controllers have almost no bias in steady-state, regardless of how the control signal is estimated, because the integral term drives the average error to zero.
When the terrain is smooth and the robots' reactions match the MPC predictions, the error variance and maxima are similar between the MPC controllers and the direct measurement of the leader. 
Only on the Hillside Loop does PI-Reflec lead to meaningfully smaller standard deviation of 4.2 cm vs 5.6 cm for the D-MPC.
Investigating the locations along the path where the largest errors occur, we observe that tight corners and changes in slope lead to errors for the MPCs. 
Tight corners challenge all controllers, as the response of the vehicle becomes less ideal close to its mechanical limits.
More interestingly, the 2D projection of the target path for the MPC leads to a discrepancy between the predicted location of the leader and the true path.
This is most visible when the leader is about to crest a hill.
The MPC prediction will continue to climb in the local plane of the leader, even though during that rollout, the slope ends and the robot will return to a flatter orientation.
The follower selects target positions to maintain the Euclidean distance to the leader's rollout but executing these commands leads to the robots becoming too close together as the follower thinks the leader will fly above the terrain. 
In the future, considering the local shape of the path along the rollout might help to mitigate this issue. 

The two MPCs result in a consistent separation bias of approximately 3 cm too close.
To investigate the cause of this bias, the D-MPC test on the Dome Loop was duplicated with the convoying order of the robots switched. 
The results of this investigation are presented in Table \ref{tab:order_switch}.
We find that the bias has the opposite sign in this configuration. 
In the default configuration, with Robot A as the leader, the bias is -2.7 cm; however, when Robot A is the follower the bias is +2.2 cm. 
This points to a hardware-specific bias from one or both robots. 
Practically, these biases are constant enough throughout the different testing environments that a constant offset could be applied to correct it for a fixed pair of vehicles.

\newcommand{\loopTableDirection}{%
	\scriptsize
	\begin{tabularx}{\linewidth}{p{0.3cm}|X|q{2cm}|q{2cm}|}
		\cline{3-4}
		\multicolumn{2}{c|}{} & Robot A leads & Robot B leads \\ \clineB{1-4}{3}
		\multicolumn{1}{|p{0.3cm}|}{\multirow{4}{*}{\rotatebox[origin=r]{90}{\parbox[c]{12mm}{\centering \vspace{-0.1cm} \textbf{Euclidean \linebreak Distance }}}}} 
		& Mean Error & -2.7 & 2.2 \\ \cline{2-4}
		\multicolumn{1}{|p{0.3cm}|}{}&RMSE & 5.1 & 3.9 \\ \cline{2-4}
		\multicolumn{1}{|p{0.3cm}|}{}&Standard Deviation & 4.3 & 3.2 \\ \cline{2-4}
		\multicolumn{1}{|p{0.3cm}|}{}&Max Error & -17.3 & 17.5 \\ \clineB{1-4}{3}
		\multicolumn{1}{|p{0.3cm}|}{\multirow{4}{*}{\rotatebox[origin=r]{90}{\parbox[c]{12mm}{\centering \vspace{-0.1cm} \textbf{Path \linebreak Tracking}}}}}
		& Leader RMSE & 2.8 & 4.4 \\ \cline{2-4}
		\multicolumn{1}{|p{0.3cm}|}{}&Leader Max Error & 9.8 & 9.8 \\ \cline{2-4}
		\multicolumn{1}{|p{0.3cm}|}{}&Follower RMSE & 4.5 & 2.9 \\ \cline{2-4}
		\multicolumn{1}{|p{0.3cm}|}{}&Follower Max Error & 10.3 & 10.5 \\ \clineB{1-4}{3}
	\end{tabularx}
	\normalsize
	\vspace{1pt}
}

\begin{figure}[t]
	\centering
	\noindent
	\includegraphics[width=\linewidth]{"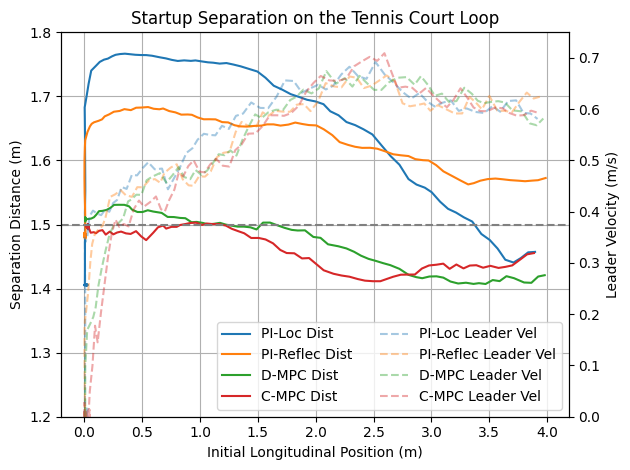"}
	\vspace{-5mm}
	\caption{Startup separation as the robots start to move. The MPC controllers maintain close to the target of 1.5 m from the initial motion. The lighter colours show the velocity of the leader as it accelerates from rest.}
	\label{fig:transient_response}
	\vspace{-5mm}
\end{figure}

\begin{table}[b]
	\caption{The effect of robot order on performance. Tested on the Dome Loop using the D-MPC. All values are in cm.}
	\centering
	\begin{tabularx}{\linewidth}{c}
		\hspace{-3mm}
		\loopTableDirection
	\end{tabularx}
	\label{tab:order_switch}
\end{table}

In Table \ref{tab:order_switch}, the RMS path-tracking error of Robot A is 2.8 cm and 2.9 cm as leader and follower, respectively.
Similarly, Robot B has RMS errors of 4.4 cm and 4.5 cm as leader and follower.
This suggests that the individual tuning of the single-robot path tracking is the primary effect rather than the role of leader or follower. 
This makes sense, as the reference path of the follower is always on the taught path regardless of the leader's position. 
This suggests that the following methods do not reduce the quality of path tracking while in a convoy.

The MPCs perform better while the leader accelerates than the PI controllers. 
For example, when the convoy starts to move, the future knowledge of the leader's state allows the follower to match the acceleration profile and maintain the target better than the reactive controllers. 
\autoref{fig:transient_response} shows the Euclidean distance over the first 4 m of forward travel on the flat Tennis Loop. 
D-MPC maintains the target most accurately during acceleration while the reactive controllers causes the follower to lag behind for approximately 3 m of transient behaviour. 

\section{LONGER CONVOY RESULTS}
\label{sec:hunter_qual_results}

\begin{figure}
	\centering\includegraphics[width=0.48\textwidth]{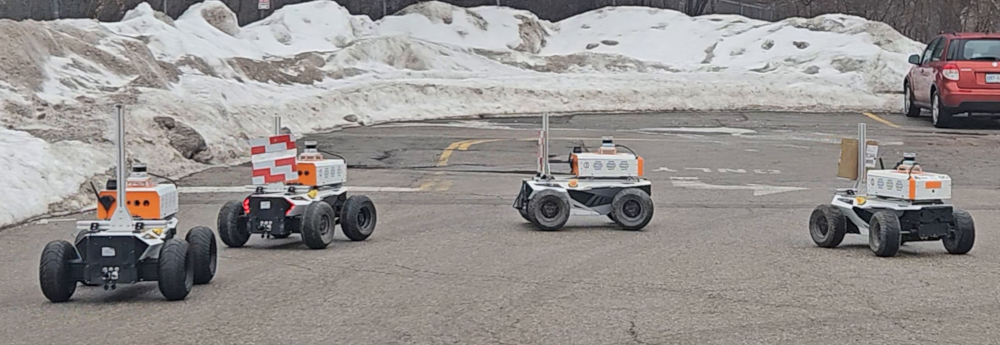}
	\caption{A convoy of four Hunter 2.0 vehicles driving using the D-MPC.}
	\label{fig:4voy}
	\vspace{-5mm}
\end{figure}

A primary advantage of using a distributed MPC instead of a centralized one is the flexibility with respect to the composition of the multi-robot system. 
For a centralized MPC, the order and kinematic model must be known ahead of time, whereas in the distributed case, robots may be added or removed from the convoy easily.
We demonstrate this flexibility by operating a convoy of four Hunter 2.0 vehicles as shown in \autoref{fig:4voy}.
The only reconfiguration of note was the networking to allow the sharing of MPC rollouts to the new vehicles. 
In our formulation of the distributed MPC, subsequent followers can subscribe to the rollout of the preceding robot, in what we refer to as a chain of followers, or the leader at the front of the convoy, referred to as a single-leader configuration.

In Table \ref{tab:4convoy}, we validate the performance of the two best performing methods, D-MPC and PI-Reflec, on a four-robot convoy. 
We evaluate D-MPC using a chain of followers configuration and a single-leader configuration. 
Due to the line-of-sight requirement with PI-Reflec, it cannot generally be run in the latter configuration, and so is evaluated with only the chain of followers mode. 
We note that, due to a transient error affecting the PI-Reflec performance of one robot, a different robot ordering was utilized for the PI-Reflec tests, placing the affected robot in the leader position.
The performance in this robot order is not substantially different from the same loop in Section \ref{sec:hunter_quant_results}, and is sufficient for comparison.
	
Interestingly, the performance of the D-MPC using a chain of followers exceeds that of the single-leader configuration. 
We also note that the PI-Reflec, while sporting the lowest RMSEs, also incurs the highest max error. 
These errors propagate down the chain, with each successive robot incurring a higher max error than its leader. 
This behaviour is not exhibited by the D-MPC methods, as they are able to act proactively according to their leader's planned actions, instead of purely reactively as PI-Reflec does.
This phenomenon has been studied under the moniker of string stability~\cite{swaroop_1996}.
The propagation and amplification of maximum errors implies a lack of string stability for PI-Reflec. 
This can cause larger issues when considering arbitrarily long convoys.

\newcommand{\loopTableFourConvoy}{%
	\scriptsize
	\begin{tabularx}{\linewidth}{p{0.5cm}|c|q{1.0cm}|q{1.6cm}|q{1.cm}|}
		\cline{3-5}
		\multicolumn{2}{c|}{} & D-MPC (Chain) & D-MPC    (Single-Leader) & PI-Reflec (Chain) \\ \clineB{1-5}{3}
		\multicolumn{1}{|p{0.5cm}|}{\multirow{4}{*}{\rotatebox[origin=r]{90}{\parbox[c]{12mm}{\centering \vspace{-0.1cm} \textbf{Robot B Euclidean \linebreak Distance}}}}}
		& Mean Error          & 1.8  & 2.7  & \textbf{0.4}  \\ \cline{2-5}
		\multicolumn{1}{|p{0.5cm}|}{}& RMSE               & 3.9  & 4.9  & \textbf{3.3}  \\ \cline{2-5}
		\multicolumn{1}{|p{0.5cm}|}{}& Standard Deviation  & 3.5  & 4.1  & \textbf{3.2}  \\ \cline{2-5}
		\multicolumn{1}{|p{0.5cm}|}{}& Max Error           & 20.2 & \textbf{17.3} & 21.2 \\ \clineB{1-5}{3}
		\multicolumn{1}{|p{0.5cm}|}{\multirow{4}{*}{\rotatebox[origin=r]{90}{\parbox[c]{12mm}{\centering \vspace{-0.1cm} \textbf{Robot C Euclidean \linebreak Distance}}}}}
		& Mean Error          & -4.0  & -5.3  & 0.4  \\ \cline{2-5}
		\multicolumn{1}{|p{0.5cm}|}{}& RMSE               & 6.2   & 5.8   & \textbf{3.6}  \\ \cline{2-5}
		\multicolumn{1}{|p{0.5cm}|}{}& Standard Deviation  & \textbf{3.2}   & 4.2   & 3.6  \\ \cline{2-5}
		\multicolumn{1}{|p{0.5cm}|}{}& Max Error           & \textbf{-16.4} & -19.3 & 23.2 \\ \clineB{1-5}{3}
		\multicolumn{1}{|p{0.5cm}|}{\multirow{4}{*}{\rotatebox[origin=r]{90}{\parbox[c]{12mm}{\centering \vspace{-0.1cm} \textbf{Robot D Euclidean \linebreak Distance}}}}}
		& Mean Error          & -0.5  & -2.4  & \textbf{0.3}  \\ \cline{2-5}
		\multicolumn{1}{|p{0.5cm}|}{}& RMSE               & \textbf{2.9}   & 5.0   & 3.8  \\ \cline{2-5}
		\multicolumn{1}{|p{0.5cm}|}{}& Standard Deviation  & \textbf{2.3}   & 4.4   & 3.8  \\ \cline{2-5}
		\multicolumn{1}{|p{0.5cm}|}{}& Max Error           & \textbf{-10.0} & -18.7 & 27.4 \\ \clineB{1-5}{3}
	\end{tabularx}
	\normalsize
	\vspace{1pt}
}

\begin{table}[b]
	\caption{Evaluation of Four Robot Convoying for PI-Reflec and two D-MPC configurations. Tested on the Mars Dome Loop. All values are in cm.}
	\centering
	\begin{tabularx}{\linewidth}{c}
	    \hspace{-3mm}
		\loopTableFourConvoy
	\end{tabularx}
	\label{tab:4convoy}
\end{table}

The Euclidean-distance reference-pose planning method can be easily replaced with one using the arclength between the robots. 
This represents a more common large-convoy setup where each robot is a common arclength distance from each other \cite{vasseur_2004}. 
The supplementary video shows a comparison around sharp corners of the Euclidean-distance and arclength modes.
Sharp turns lead to the most significant behavioural change. 
An arclength implementation is difficult with a direct measurement of the leader's position.

The D-MPC and PI-Reflec controllers perform more reliably than C-MPC and PI-Loc. 
Additionally, due to their distributed nature, both are scalable to longer convoy chains. 
On rapid elevation changes, using a direct measurement of the leader's relative position could be advantageous.
However, the D-MPC approach provides more flexibility as the robots can convoy beyond line-of-sight and longer convoys are unaffected by disturbances along the chain. 
The D-MPC makes the fewest assumptions about the operating environment and geometric arrangement of the convoy, which makes it widely applicable.
For this reason, we selected the D-MPC for evaluation in lunar-analogue cargo-transport experiments.

\section{NETWORK RESULTS}

As the D-MPC relies on inter-robot communication, we collect statistics on the communications used during operation, and the effects of network traffic on control performance.
The Hunters are each outfitted with a MediaTek MT7621A WiFi SoC and a MediaTek MT7603E WiFi single chip, configured to support a bitrate of 65 Mbit/s. 
In the following statistics, Robot B is configured as the central node in the network topology. 
ROS2 timestamps are used to measure latency, and tcpdump is used to capture statistics on network traffic.
We evaluate the bandwidth of the communication stream in a two-robot convoy.
To evaluate the scalability of our method, we collect the same information on a four-robot convoy scenario.
All network experiments are run with a target separation of 2.5 m between successive robots. 

\begin{figure}
	\centering
	\noindent
	\includegraphics[width=0.85\linewidth]{"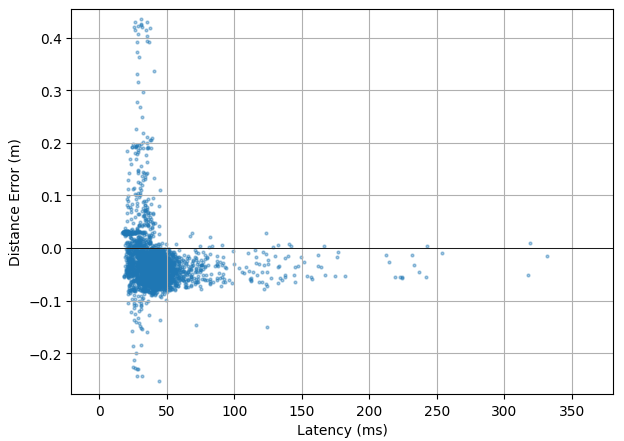"}
	\caption{There is no correlation between the network latency and the Euclidean distances between the robots. The outlier errors predominately occur with low latency.}
	\label{fig:net-lat}
	\vspace{-3mm}
\end{figure}

During our two-robot experiments, the bandwidth occupied by messages necessary for convoy operation is measured as 13.83 KB/s.
A mean latency of 33.67 ms for these messages was measured. 
The primary operational concern is the effect of network outages and transient effects on inter-robot distance. 
In Figure \ref{fig:net-lat}, we plot the error against the corresponding measured latency. 
We note a negligible relationship between the two variables. 
As the D-MPC uses extrapolation to account for communications latency and dropouts, it is robust to minor variations in message timing, as long as the MPC rollout is accurate regarding the future actions of the leading robot.

To measure the effects of increased network traffic on required hardware, we analyze the same statistics with a four-robot convoy.
We note negligible changes in message bandwidth usage, measuring 13.75 KB/s, 13.70 KB/s, and 13.94 KB/s for Robots B, C, and D in the convoy, respectively. 
Latency suffers from increased traffic and increases with the distance from the central network node. 
The mean measured latency values are 46.06 ms, 51.93 ms, and 57.90 ms for Robots B, C, and D, respectively.

\section{LUNAR ANALOGUE TESTING SETUP}
As part of the \ac{LUV} development, this field test provided an opportunity to simulate lunar cargo transport.
The proposed mission concept requires the cargo to be equipped with actuated legs that can lift it clear of the rovers' cargo decks and lower itself onto the vehicles. 
This leads to three different mission stages: cargo pickup, convoy transport, and habitat alignment. 
During cargo pickup, one or both rovers are controlled remotely to define a safe teach path that travels from the habitat to the cargo's landing location. 
We anticipate that this network of paths will be reused for multiple deliveries, which speeds up the operation. 
The vehicles can follow this route individually or in an uncoupled convoy until they are near the cargo. 

\begin{figure}
  \includegraphics[width=0.95\linewidth]{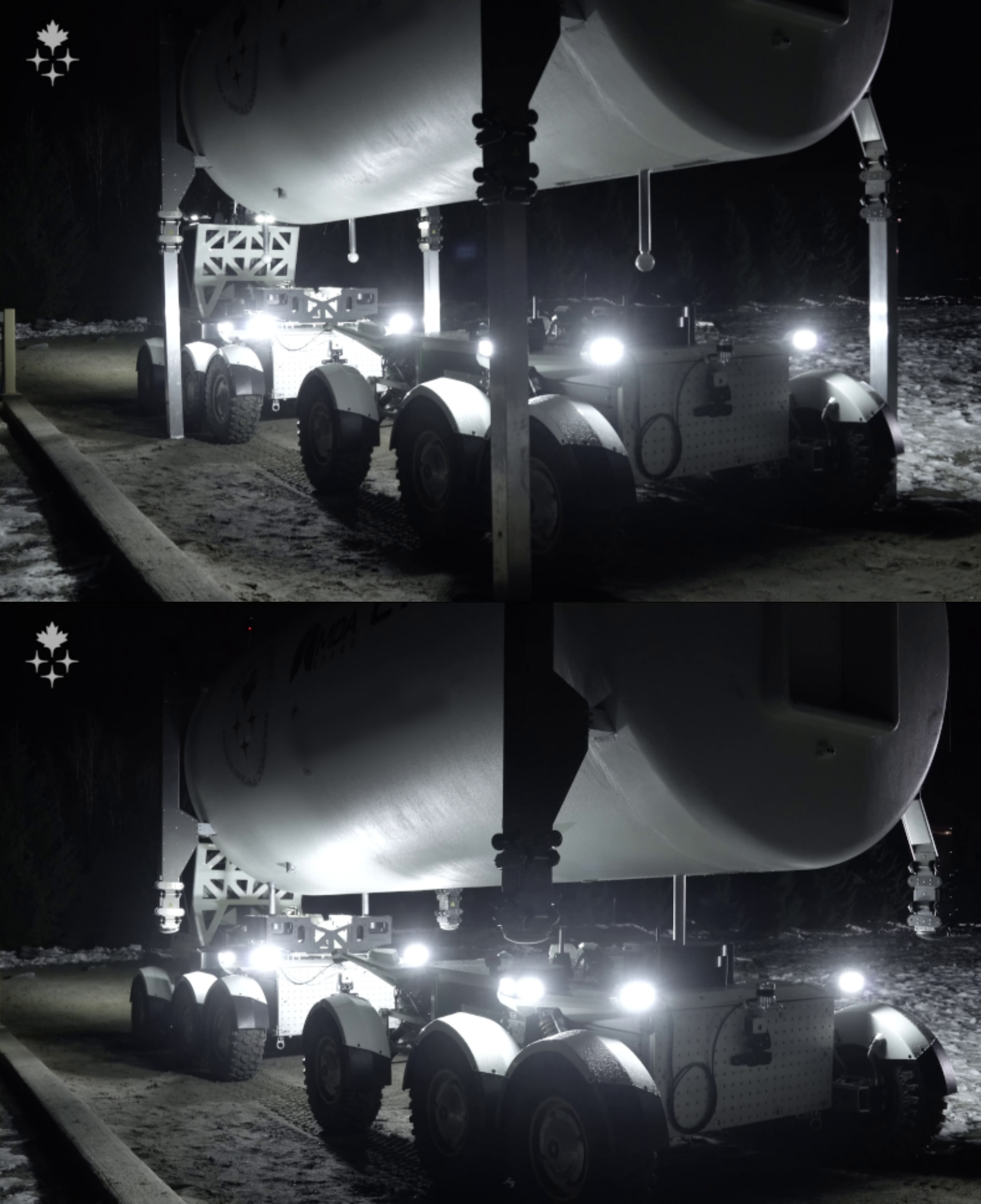}
  \caption{The cargo as it was collected during the \textit{Night} run. The legs extend to raise the cargo above the deck of Rover B so that it can drive completed underneath. The legs retract to lower the cargo onto the rovers. The hold-down mechanism allows for about 15 cm of position error for each rover during initial alignment. Photo Credit: CSA}
  \label{fig:legs-retract}
  \vspace{-6mm}
\end{figure}

With the cargo legs extended, one rover can fit completely underneath the cargo.
As it navigates under, it continues teaching a new branch specific to the current piece of cargo.
Once it is positioned for pickup, that map is sent to the other rover to expedite the alignment process. 
Once aligned, the cargo legs retract to lower the cargo onto the vehicles and it is secured into place. 
Once the cargo is secured, a coupled convoy begins to bring the cargo back to the habitat. 
\autoref{fig:legs-retract} shows an example of cargo loading with both rovers already aligned to the cargo for pickup. 
We demonstrate the ability of the convoy to switch direction smoothly while the cargo is connected, which is likely to be required along the route back to the habitat.
The final step is a reverse-docking manoeuvre  where the cargo is backed up into alignment with the existing habitat structure. 
The precision of this step is critical to allow for a docking or airlock to connect the new module to the existing structure. 
At this stage, the cargo legs can extend to allow the rovers to drive out from underneath the cargo.

\subsection{HARDWARE}
\begin{figure}
  \includegraphics[width=0.98\linewidth]{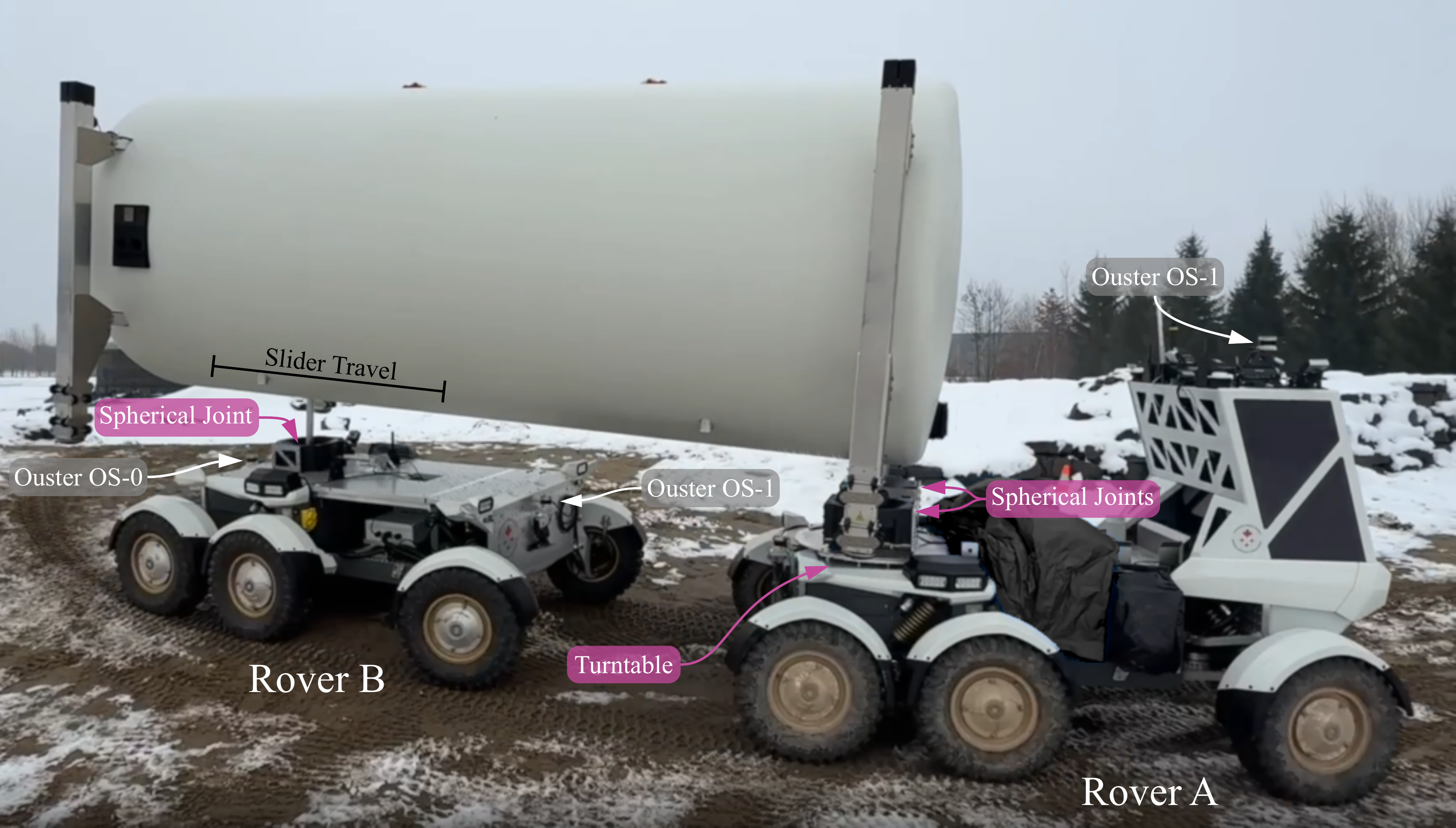}
  \caption{The two \lelr \ rovers carrying the cargo. The sensors used on each vehicle are highlighted in callouts. The cargo joints are highlighted with pink callouts on both robots. The slider is inside of the cargo but its range is illustrated.}
  \label{fig:convoy}
  \vspace{-5mm}
\end{figure}

Two Lunar Exploration Light Rovers (\lelr s) \cite{bakambu_2016} are used. 
The base platform is the same, but they each have been independently modified. 
The convoy is shown in \autoref{fig:convoy} with Rover A on the right and Rover B under the cargo on the left.

Each \lelr \ is six-wheel drive with two independently steered wheels controlled by electric actuators and four fixed wheels.
The steering replicates an Ackermann geometry virtually.
The low-level motion controller was modified based on \cite{chhabra_2016} to convert linear velocity commands into the torques for the six wheels and a central Ackermann steering angle to the angles for each steering actuator.

The rubber tires are 76.3 cm in diameter, allowing obstacles up to 30 cm high to be cleared. 
\LELR \ is powered by a set of lithium-ion batteries with a total capacity of 12 kWh at 96V.
Each vehicle has an unloaded mass of 800 kg and can carry 300 kg at a maximum velocity of 13 km/h.

\subsection{Cargo}

The cargo is a 5.475 m long cylinder with an internal steel frame and a polystyrene exterior. 
It has a mass of 475 kg.
Three steel columns with ultra-high molecular weight (UHMW) plastic spheres provide the support interfaces for the cargo. 
The rear column attaches to a sliding carriage that allows it to move $\pm$ 50 cm from the nominal position. 

Mounted to the deck of both rovers is the cargo hold-down mechanism.
The hold down on Rover A consists of two spherical joint receptacles mounted on top of a turntable. 
This compound joint constrains the position of the front of the cargo and the roll of the cargo.
The hold down on Rover B consists of a single spherical joint receptacle.
However, the rear support post on the cargo is connected to a prismatic joint. 
This compound joint constrains the vertical position of cargo's rear as well as the lateral position of the rear with respect to Rover B. 
These six kinematic constraints fully define the motion of the cargo but leave the two rovers kinematically independent from each other.

\subsection{Communication}

Each subsystem: Rover A, Rover B, the cargo, and the \ac{GCS}, has its own Pep-Wave Max BR1 router managing a subnetwork. 
Rover A is used as the time source for \ac{NTP} for synchronizing the clocks. 
Rover A is the default hub, that the cargo, Rover B, and the \ac{GCS} each connect to with 2.4 GHz WiFi. 
A future mesh network could be used to make inter-system communication more resilient. 

Communication between Rovers A and B is required for convoy motion. 
No other time-sensitive communication paths exist. 
13.83 KB/s of data is transferred between the rovers for convoying.
We observed 166.67 KB/s wire transfer over the ROS2 Zenoh port while driving. 
Topometric map transfer requires more data but can be completed while the rovers are stationary. 

\subsection{Sensors}
The fields of view of the navigation lidar sensors on both vehicles are a critical piece of information for convoy navigation. 
The sensors used for convoy navigation are illustrated in \autoref{fig:convoy}.
Rover A has one lidar: an Ouster OS-1 mounted parallel to the ground on Rover A's cab.
There are two additional lidars on the platform but they are unused.
Rover B has two lidars: an Ouster OS-1 mounted on the front face looking forward and an Ouster OS-0 mounted on the rear face looking backwards. 
All lidars are mounted along the midline of the vehicles. 
Rover A's OS-1 has a 360$^\circ$ view of the environment. 
With the cargo loaded, it is reduced to 270$^\circ$.
On Rover B, the two lidars each have a 180$^\circ$ view at the front and rear when is operates independently. 
When driving in a convoy, approximately 120$^\circ$ are blocked by Rover A. 
These two lidars are stitched together, accounting for their known relative positions, to create a 360$^\circ$ or 240$^\circ$ field of view. 
To minimize the lag introduced by waiting for both point clouds to merge, an initialization step looks at the relative time offset between the point clouds.
Triggering the merge on the correct scan guarantees a maximum lag of half a sensor period, in this case 50 ms. 
The per-point timestamps are retained for downstream motion compensation. 
Using a hardware trigger for all lidars on one vehicle would eliminate this problem in the future.  

To evaluate the convoy's performance, we have additional sensors to assess the navigation, Euclidean distance accuracy, and habitat alignment. 
A Novatel SPAN \ac{INS} captures the position and orientation of Rover A to evaluate the ground-truth performance of the navigation and localization. 
In the same manner as the Hunter 2.0 ground-truth, a retro-reflective panel is mounted to the rear of Rover A and is tracked using the front lidar on Rover B.
When the cargo is loaded, a linear potentiometer in the prismatic joint captures the true Euclidean distance between the vehicles; this is the ground-truth measurement of the convoy inter-robot distance. 
Finally, a cargo-mounted ultrasonic sensor measures the distance between the rear of the cargo and the habitat docking surface. 

\begin{table*}[t]
    \centering
\caption{Convoy results on the CSA Analogue Terrain Route. All error values in cm.}
\label{tab:csa_results}
\begin{tabular}{lr|rr|rr|rrrr|}
\cline{3-10}
 &
  \multicolumn{1}{l|}{} &
  \multicolumn{2}{c|}{\begin{tabular}[c]{@{}c@{}}\textbf{Path Tracking Error}\\ \textbf{Rover A}\end{tabular}} &
  \multicolumn{2}{c|}{\begin{tabular}[c]{@{}c@{}}\textbf{Path Tracking Error}\\ \textbf{Rover B}\end{tabular}} &
  \multicolumn{4}{c|}{\textbf{Euclidean Distance}} \\ \cline{2-10} 
\multicolumn{1}{l|}{} &
  \multicolumn{1}{l|}{Path Length (m)} &
  \multicolumn{1}{c|}{RMSE} &
  \multicolumn{1}{c|}{Max} &
  \multicolumn{1}{c|}{RMSE} &
  \multicolumn{1}{c|}{Max} &
  \multicolumn{1}{c|}{Mean} &
  \multicolumn{1}{c|}{RMSE} &
  \multicolumn{1}{c|}{Min} &
  \multicolumn{1}{l|}{Max} \\ \hline
\multicolumn{1}{|l|}{Morning} &
  333.2 &
  \multicolumn{1}{r|}{6.9} &
  26.4 &
  \multicolumn{1}{r|}{8.2} &
  46.8 &
  \multicolumn{1}{r|}{-2.2} &
  \multicolumn{1}{r|}{8.8} &
  \multicolumn{1}{r|}{-32.6} &
  21.7 \\ \hline
\multicolumn{1}{|l|}{Noon} &
  347.2 &
  \multicolumn{1}{r|}{6.6} &
  30.5 &
  \multicolumn{1}{r|}{8.9} &
  54.8 &
  \multicolumn{1}{r|}{-1.6} &
  \multicolumn{1}{r|}{9.0} &
  \multicolumn{1}{r|}{-33.4} &
  22.1 \\ \hline
\multicolumn{1}{|l|}{Night} &
  329.8 &
  \multicolumn{1}{r|}{6.9} &
  24.6 &
  \multicolumn{1}{r|}{8.6} &
  45.9 &
  \multicolumn{1}{r|}{-0.1} &
  \multicolumn{1}{r|}{9.8} &
  \multicolumn{1}{r|}{-33.0} &
  28.6 \\ \hline
\multicolumn{1}{|l|}{Forward} &
  1122.2 &
  \multicolumn{1}{r|}{7.0} &
  30.5 &
  \multicolumn{1}{r|}{8.4} &
  54.8 &
  \multicolumn{1}{r|}{-1.7} &
  \multicolumn{1}{r|}{9.1} &
  \multicolumn{1}{r|}{-33.4} &
  22.1 \\ \hline
\multicolumn{1}{|l|}{Reverse} &
  219.6 &
  \multicolumn{1}{r|}{4.1} &
  26.4 &
  \multicolumn{1}{r|}{5.6} &
  26.7 &
  \multicolumn{1}{r|}{-0.2} &
  \multicolumn{1}{r|}{9.5} &
  \multicolumn{1}{r|}{-28.3} &
  \multicolumn{1}{l|}{28.6} \\ \hline
\multicolumn{1}{|l|}{Total} &
  1341.8 &
  \multicolumn{1}{r|}{6.6} &
  30.5 &
  \multicolumn{1}{r|}{8.0} &
  54.8 &
  \multicolumn{1}{r|}{-1.4} &
  \multicolumn{1}{r|}{9.2} &
  \multicolumn{1}{r|}{-33.4} &
  \multicolumn{1}{l|}{28.6} \\ \hline
\end{tabular}
\end{table*}

\section{RESULTS}
Ultimately, the goal of evaluating the D-MPC on representative hardware is to constrain the requirements for cargo couplings.
The slider and turntable both have large margins during development, but the smaller those can be, the better. 
For cargo transport, we are interested in the distribution of Euclidean inter-robot distance errors.
Although the cargo should be tolerant to reaching the hard stops, ideally no joint will reach its mechanical limits.

In addition to the Euclidean distance, it is important that both vehicles continue to follow the taught path accurately. 
One of the core benefits of repeating paths during cargo transport is to reduce the amount of local terrain assessment that is required while in motion. 
The more precisely the rovers follow in their tracks, the safer the repeat will be. 
This is especially important for the following vehicle as its forward view of the terrain is obstructed by the lead vehicle and the cargo itself. 
We evaluate the path tracking of both vehicles on the same path and compare that performance to convoy runs without the cargo loaded onto the vehicles. 

Additionally, we evaluate the lidar localization performance on Rover A. 
Due to the presence of the cargo, an \ac{INS} was not installed on Rover B.
This prevents the evaluation of localization on that vehicle. 
This data helps position the single-rover performance of \lelr \ with other systems running teach and repeat.  

At the CSA Analogue Terrain, we chose a path, shown in \autoref{fig:eval_paths}(b) that makes use of most of the size of the terrain and drives over \textit{Saddle Mountain}, a hill with a maximum approach pitch of 7.8$^\circ$.
The path contains a central figure eight with two branches for pickup and drop off. 
In the evaluation runs, we follow the figure eight twice to increase the path length for evaluation. 
At the end of one branch is a mock habitat structure that is used to evaluate alignment quality. 
The nominal forward speed for \lelr \ is 0.5 m/s, with the MPC constrained to select speeds less than 0.75 m/s.

For the final alignment, we use the rear-facing distance sensor on the cargo to measure the distance between the cargo door and the habitat structure. 
Angular misalignment is important but is difficult to quantify with the current data. 
We report the angular misalignment of the cargo with respect to the target path but this does not include misalignment between the habitat and the target path itself.

\begin{figure*}[p]
  \centering
  \includegraphics[width=\textwidth]{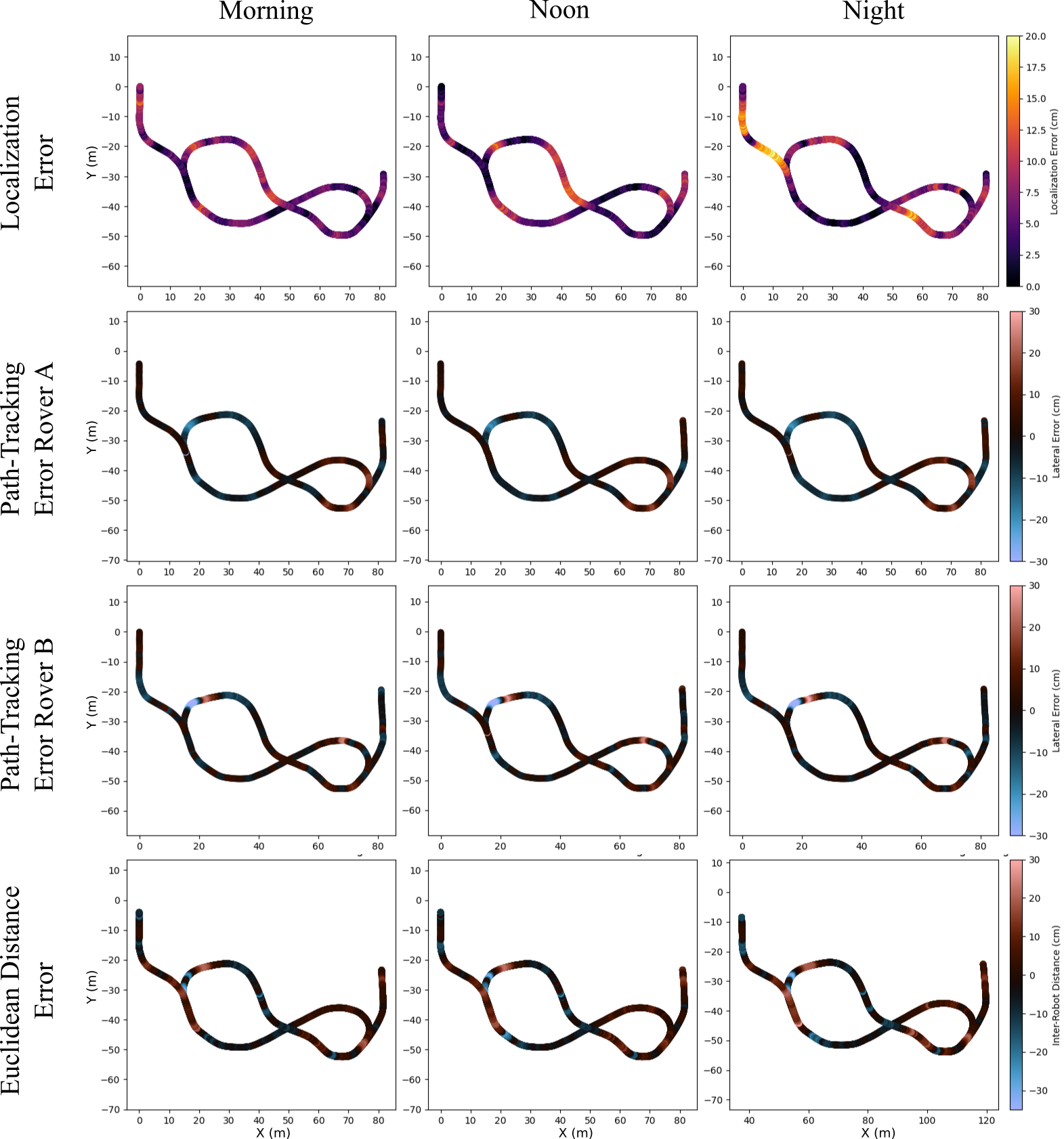}
  \caption{A 4$\times$3 grid of the localization, path-tracking, and Euclidean distance errors on each of the three evaluations. The path-tracking and Euclidean distance errors are consistent between runs. Localization errors are consistent in the \textit{Morning} and \textit{Noon} runs but change in the \textit{Evening}.}
  \label{fig:csa_loop_plots}
\end{figure*}

\subsection{Convoy Driving}

\autoref{tab:csa_results} summarizes the core results of the convoy driving. 
The Euclidean distance and path-tracking error are broken out for each convoy run and aggregated by driving direction and total. 
The sign of the path tracking error is defined by the local path frame with positive errors driving to left of the path as taught. 
There were three repeats of the Analogue Terrain Loop and two reset repeats that moved the cargo from the habitat back to the start to test again. 
In \autoref{fig:csa_loop_plots}, the top row illustrates the magnitude of the localization error of Rover A, the middle two rows show the path-tracking error of each rover, and the bottom row shows the Euclidean distance error assigned to the position of Rover A.

To evaluate the localization error, post-processed \ac{GNSS}-\ac{INS} data was compared between the teach and repeat paths of the vehicle. 
The localization and mapping timestamps were used to select the poses to compare.
That difference was compared to the lidar localization estimate.
Over the course of the day, Rover A had an overall \ac{RMSE} of 9.5 cm.
Examining the first row of \autoref{fig:csa_loop_plots}, the errors are correlated with position. 
It is notable that the first two runs have similar spatial error distributions while the \textit{Night} run has a distinct distribution.
Although post-processing of the \ac{GNSS} was used, the larger time gap could suggest that the visible \ac{GNSS} satellites impacted the results. 

\begin{figure}
  \centering
  \includegraphics[width=0.9\linewidth]{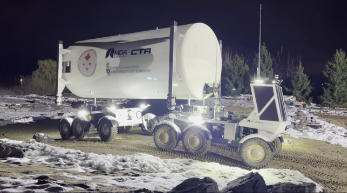}
  \caption{Rovers A and B as they crest \textit{Saddle Mountain} during the \textit{Night} run. The local planes that each rover drives in are the least aligned in this location. This is the leading theory for the maximal Euclidean distance errors in this location.}
  \label{fig:saddle}
\end{figure}

Looking at the Euclidean distance error, the bias is small suggesting that the two rovers are time-synchronized well.
The maximum and minimum errors, while much larger than the Hunter vehicles, do not approach the mechanical limits at $\pm$50 cm. 
The total range exercised is 62\% of the available range. 
The fourth row of \autoref{fig:csa_loop_plots} illustrates that the maximum inter-robot distance errors occur in the same spot on the path each time. 
This error correlates with the blue sections in the fourth row where Rover B is too close to Rover A these maximum errors reach as high as 54.8 cm. 
This location appears to have errors in all of localization, path-tracking, and Euclidean distance tracking. 
As the rovers crest \textit{Saddle Mountain} (\autoref{fig:saddle}), they veer wide of the path in every repeat. 
Interestingly, Rover B has an extra lateral deviation earlier in the path not exhibited by Rover A (the orange region to the right of the common blue deviation).
Although no path-tracking performance differential was observed on the Hunter vehicles, it is possible that the follower reference pose selection when the leader is tracking poorly causes unexpected deviations for the follower. 
We investigate the cross-correlation of the estimation error of the Euclidean distance and the observed Euclidean error.
The \textit{Morning, Noon,} and \textit{Evening} runs show similar patterns. 
The actual error lags estimation error by 1.6 s, 1.7 s, and 2.5 s, respectively. 
With correlation coefficients of 0.52, 0.56, and 0.71 respectively, the Euclidean distance errors do not come solely from estimation errors but appear large enough to have an effect. 
Given the significant inertia of the systems and the fact that the estimation errors exhibit a high autocorrelation, it appears that erroneous position estimates on either vehicle manifest in tracking errors. 
Over the evaluation repeats, Rover B has a larger path-tracking error than Rover A with 8.0 cm vs 6.6 cm.
It is notable that both rovers have smaller path-tracking errors while reversing. 
The total distance driven in reverse is approximately 20\% of the total meaning that a direct comparison between these two modes is inconclusive.

\subsection{Cargo Alignment}

\begin{figure}
  \centering
  \includegraphics[width=0.9\linewidth]{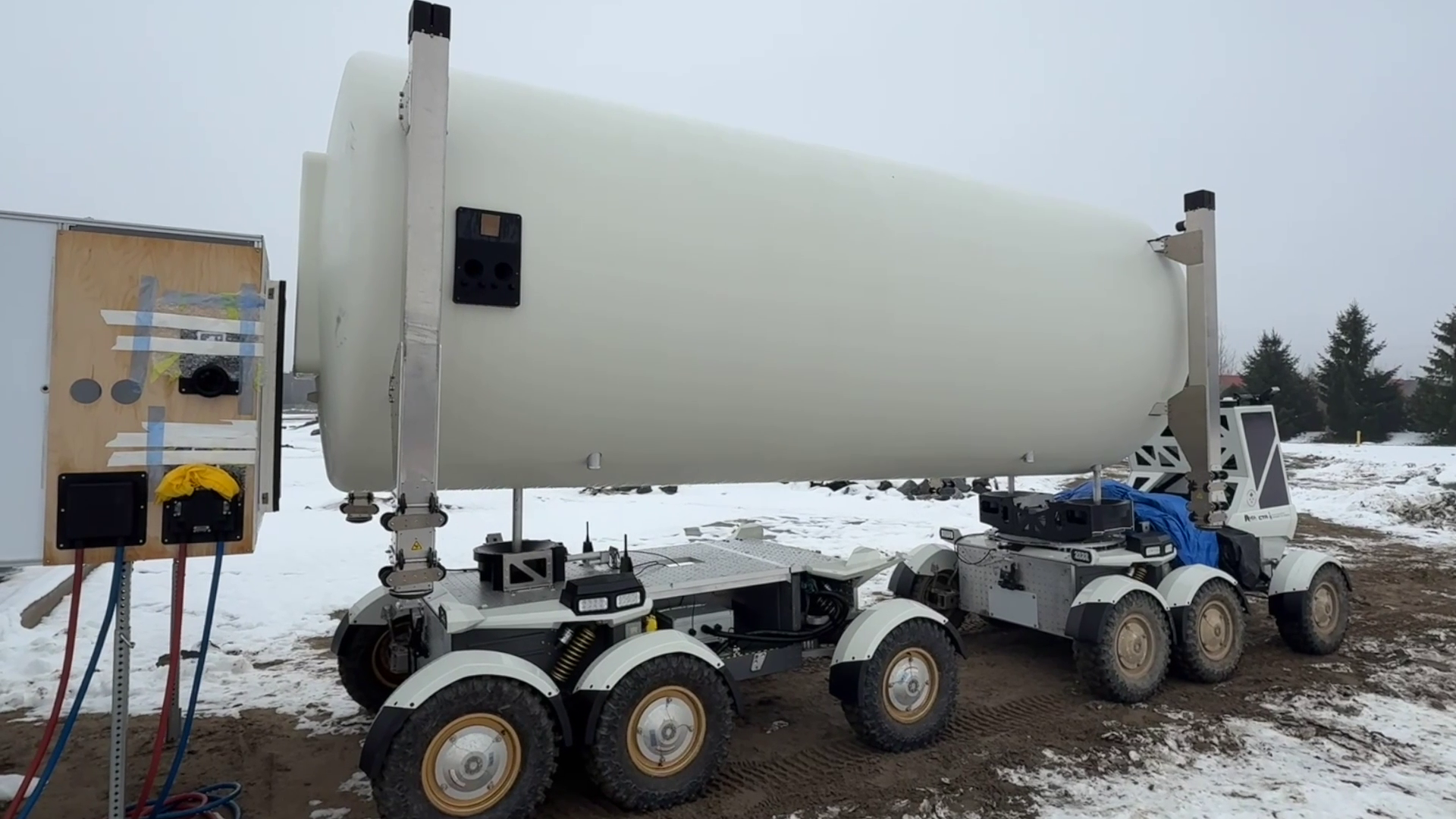}
  \caption{The cargo in the final alignment relative to the habitat in the \textit{Morning} run. The habitat is 7.9 cm from the rear of the cargo. Once the rovers drive away, the cargo will lower to the same height as the habitat door.}
  \label{fig:habitat}
\end{figure}

The reverse alignment into the habitat is an important step for mission success.
Three alignments in \autoref{tab:cargo_dist} correspond to the repeats above, and one additional test was performed for more data. 
The average clearance between the cargo and habitat doors is 11.5 cm with a 2.8 cm standard deviation. 
Qualitatively, the lateral error and yaw offset are not visible relative to the size of the cargo module. 
\autoref{tab:cargo_dist} reports the heading and lateral alignment error of the rear door of the cargo with respect to the taught path. 
There is an additional alignment error of the taught path to the habitat. 
\autoref{fig:habitat} shows the cargo in position relative to the habitat. 

\begin{table}[b]
    \centering
    \caption{The clearance between the cargo door and the mock habitat.}
    \scriptsize
    \begin{tabularx}{\linewidth}{|C|C|C|C|}
        \hline
        \textbf{Run Description} & \textbf{Door Clearance (cm)} & \textbf{Door Lateral Error (cm)} & \textbf{Cargo Heading Error ($^\circ$) } \\
        \hline
        Morning & 7.9 & 3.7 & -0.3\\
        \hline
        Noon & 9.9 & 2.4 & -0.1\\
        \hline
        Night & 15.5 & 2.0 & 0.3\\
        \hline
        Cargo Realign & 12.5 & 11.7 & -1.9\\
        \hline
    \end{tabularx}
    \label{tab:cargo_dist}
\end{table}

\subsection{Effect of Cargo on Convoying}

\begin{figure}[t]
  \centering
  \includegraphics[width=0.9\linewidth]{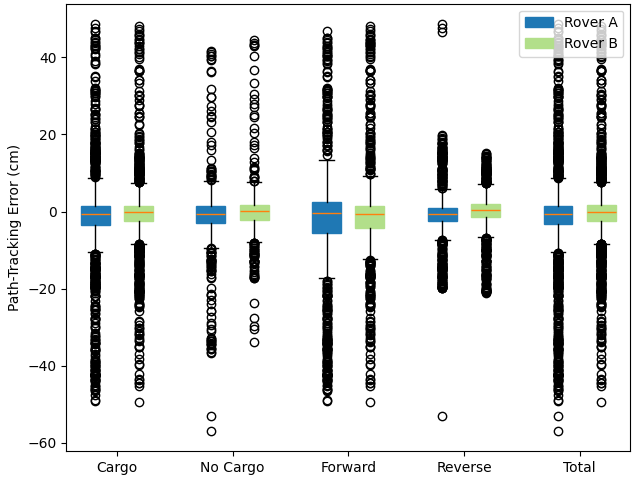}
  \caption{A comparison of path-tracking errors with and without cargo at the Sand Pit test site.}
  \label{fig:cargo_effect}
\end{figure}

Due to limited testing time at the CSA Analogue Terrain, we were unable to repeat the same route detailed in \autoref{tab:csa_results} without the cargo. 
However, in earlier testing at the Sand Pit, there is a direct comparison of the effect of the cargo weight on the system.
We run the same 155 m, shown in \autoref{fig:eval_paths}(a), in all combinations of forward and reverse, and with and without cargo. 
Looking at the results in \autoref{fig:cargo_effect}, there is a not a statistically significant effect of operating with or without the cargo.
Interestingly, there is a statically significant improvement in convoy performance when the rovers drive in reverse. 
This supports the evidence in \autoref{tab:csa_results} that the reversing errors are smaller not just because of an easier path to track.
Looking at the \textit{Total} columns Rover A has a larger inter-quartile spread and a larger \ac{RMSE} of 9.7 cm vs 7.5 cm. 
The supplementary video shows an overhead comparison of the convoy driving with and without cargo during this test. 
It is unclear if the improvement in reverse reflects better tuning of both rovers in reverse or if improved tuning on Rover B means that when it acts as the leader it is easier for Rover A to follow the path and hold the correct Euclidean distance.

\section{DISCUSSION}

While the single-rover path-tracking performance is not the key benchmark for this study, it is worth highlighting that the path-tracking error decreased to 6.6 cm from 10.1 cm and the maximum to 30.5 cm from 40.1 cm in previous studies using the same vehicle \cite{krawciw_2026}.
This is directly attributable to the change from a unicycle model to the bicycle model for the \ac{MPC}. 
An advantage of the distributed approach is that the vehicles in the convoy do not require knowledge of each other. 
If every robot publishes its MPC trajectory and follows a leader's trajectory, the details of the robot's kinematics or controller are irrelevant.
This will allow robot-specific models to be upgraded to improve local performance without directly impacting the rest of the convoy.

This field campaign provided an opportunity to explore the challenges related to teaching and repeating with disparate robots. 
While the control of both vehicles was similar, their fields of view and coupling capabilities meant that each rover was used to teach different parts of the route. 
With significant viewpoint overlap, lidar localization was successful.
It was difficult to place the lidar on Rover B to create a similar 360$^\circ$ view as on Rover A. 
With the rear lidar only, Rover B could repeat paths mapped on Rover A without cargo. 
However, once the cargo was loaded, Rover A could not localize to maps created from the rear lidar on Rover B and vice versa.
For this reason the front lidar on Rover B was mounted. 
By stitching the two lidars together, the two rovers could reliably repeat along routes taught by the other regardless of the cargo configuration. 

The wireless link between the rovers was used to send the new maps between the two rovers. 
However, an improved deterministic method of synchronizing the maps between the rovers will be required. 
Currently, the maps are shared back and forth between the two rovers but cannot resolve when a new branch is created by both rovers at the same time.  
The topological backbone of the map should make it straightforward to merge maps that have common vertices. 

It is challenging to synchronize the clocks on multiple robots.
In this work, the clocks are synchronized using \ac{NTP}, which should maintain sub 5 ms accuracy between the rovers \cite{ferrari_2020}. 
Clock asynchrony manifests as an inter-robot distance bias for the D-MPC. 
This occurs because the leader's anticipated motion is interpolated at a different time than it should be. 
During early testing, there was an error with the timestamps when the front and rear lidar point clouds were merged on Rover B. 
The effect was clear as the bias would switch sign depending on which robot was leading. 

In these testing conditions, 62 cm of range was required on the prismatic joint to prevent hard stops from limiting the performance. 
However, looking at the second and third rows of \autoref{fig:csa_loop_plots}, there is a clear spatial correlation of the path-tracking and Euclidean-distance errors.
Systematic errors provide an avenue for improvement with the existing hardware. 
Especially for path-tracking, where the rovers will traverse the same route many times, adaptive control methods, such as \cite{lenain_2007, chen_2015, zhou_2022} could be applied to reduce these errors in subsequent repeats. 
Ultimately, more experimentation will be required with cargo of different inertia to evaluate to what extent the systematic errors in Euclidean distance are connected to the specific cargo properties. 
However, given that the errors remain similar with or without the cargo, we anticipate that there is an opportunity for adaptive control gains even with various types of cargo. 

During mock mission operations, it was quickly evident that operating the rovers together as a convoy without cargo was useful.
During cargo collection, it was safer to run the rovers in a convoy than to send them sequentially from the habitat to the cargo landing position.
By operating as a convoy, a safe inter-robot distance was naturally enforced.
Additionally, keeping the rovers closer together improves the wireless connectivity between the vehicles. 

To collect the cargo, Rover B was remotely teleoperated underneath the cargo. 
While this operation was completed, Rover B mapped the local environment and recorded its motion as a teach path. 
Once it was in position for cargo collection, it would transmit the new section of the map to Rover A, which could then independently repeat that section into cargo pickup alignment. 
This was a useful feature of teach and repeat; the alignment took Rover B 30:20 teleoperated, repeating the same alignment took Rover A 5:41.
For a longer branch from the main route, this difference would be even more stark.

The cargo hold-down design means that the longitudinal position of the cargo with respect to the habitat is fully defined by Rover A. 
This is inconsistent with the leader-follower controller in this location as Rover B leads during the docking manoeuvre. 
As the follower, Rover A and the cargo stop when the leader (Rover B) commands it to. 
We anticipate that a distinct end-of-path behaviour is warranted whereby Rover B parks and Rover A continues at a very slow speed to complete the docking.
However, direct measurement of the habitat from Rover A is difficult as the cargo and Rover B occlude Rover A's view of the habitat. 

A second challenge caused by Rover A defining the longitudinal position of the cargo occurs during cargo removal from the habitat. 
When the rovers drive back underneath a docked cargo to move it (for example to  dispose of a resupply container) as the legs lower, the cargo will slide back into the neutral position of the hold down mechanism. 
If Rover A is too close to the habitat, the cargo can slide backwards into the habitat door. 
Operationally, this effect can be avoided, and a forward motion away from the habitat may even be desirable but reinforces the necessity to control the rovers differently while operating in close proximity to the habitat.

\section{LESSONS LEARNED}
Testing and evaluation of the multi-robot system required more safety layers than for a single rover to reduce the risk of damage to the vehicles. 
To minimize the risk of collision while testing, a safety monitor on both robots ran in the background. 
There are three modes that each rover could operate in: separate, convoy, and coupled. 
In both convoy and coupled, the robots drive in sync, but coupled means that the cargo is loaded and the inter-robot distance constraints are harder. 
The rovers can be soft or emergency stopped. 
While in convoy or coupled, a soft-stop or e-stop on either robot will soft-stop the other. 
These soft and emergency stops can be triggered by many conditions; however, we add two conditions to generate soft stops. 
To convoy successfully, the lead rover must communicate its predicted motion to all followers. 
Therefore, a stable communication link is required. 
A time-stamped heartbeat is sent bidirectionally between the vehicles. If the latency is greater than 200 ms or it has been more than 200 ms since the last message was received, the system will soft-stop.
Similarly, there was a risk of robot collisions due to the close-range operations. 
The lidar reflector used for ground-truth evaluation monitored the inter-robot distance at all times and based on whether the rovers were convoying or coupled would trigger a soft-stop if hard limits were nearly met.

The network setup on the path-to-flight rovers was significantly more reliable than the network on the Hunters. 
This reflected the higher-quality routers but highlighted important considerations for testing and evaluating multi-robot systems. 
From an ease of operations perspective, the reliable network connection made it easier to remotely connect to the rovers and set up experiments; however, the Hunter experience emphasized how important it was to require as little interaction as possible. 
By automating procedures for operating and removing the ground station as a central link, we successfully completed multiple convoyed repeats where communication with the ground station was lost and the convoy continued uninterrupted, until eventually the ground re-established a link. 
For mission operations, it would be beneficial to make the heartbeat link limit based on the worse-case rover constraints. 
If the rovers are coupled by a piece of cargo, there will be tight constraints on inter-robot distance; however, their proximity likely means that the communication will be reliable. 
If the rovers are uncoupled and driving separately, there will likely be a long separation that provides more leniency for communication due to the larger margin between rovers. 
Characterizing the effect of the MPC rollout update rate with convoy performance would be a good next step to establish requirements for convoy communication.

\section{CONCLUSION}

Using a distributed model predictive controller allows for large rovers to carry long and heavy pieces of cargo safely and smoothly that would be overload a single vehicle. 
The use of passive joints on the cargo hold-down mechanisms allowed the convoy to operate without requiring explicit modelling of the cargo in the MPC. 
In small-scale testing, the prismatic joint was exercised for 40 cm of the total 55 cm range on the Hillside Loop.
The prismatic joint on the large cargo was exercised for 62 cm out of the total 100 cm range. 
With further tuning of the controller, and adaptive improvements online, we anticipate that this range can be reduced further. 
From a mission perspective, lidar teach and repeat was consistently accurate enough to place the cargo close to the final position and maintain necessary clearances from the habitat. 
In future work, we will continue to refine a mission concept to dock the cargo module to the existing habitat.
The flexibility provided by powered legs was useful as it allowed cargo loading and unloading to occur in any location without external infrastructure. 

Lidar teach and repeat followed the target paths reliably with all robots tested tracking with less than 10 cm \ac{RMSE}.
On the Hillside Loop, the Hunter Robot A had a self-reported path-tracking \ac{RMSE} of 7.2 cm and Hunter Robot B 6.1 cm.
On the \lelr \ vehicles, \ac{RMSE} values of 6.6 cm and 8.0 cm on Rovers A and B respectively were observed at the Analogue Terrain. 
The results from the small-scale Hunter vehicle transferred to the \lelr \ vehicles weighing more than ten times as much.
Based on the comparison of the D-MPC to existing baselines and the practical benefits observed in the field of unifying the controller with or without cargo, this approach is a viable path forward for large-scale cargo transport for the Artemis program.

\section*{ACKNOWLEGEMENTS}
The authors thank the rest of the MDA Space rover team for their assistance supporting the rover design. 
They thank the Canadian Space Agency for use of the Analogue Terrain.
They thank Hexagon for the use of the Inertial Explorer Software. 

\printbibliography

\end{document}